\documentclass[12pt]{iopart}
\usepackage{iopams}
\usepackage{dcolumn}
\usepackage{bm}
\usepackage{graphicx}
\usepackage{framed,graphicx,color}
\usepackage{wasysym}
\usepackage{marvosym}
\usepackage{cite}

\definecolor{shadecolor}{rgb}{0.8, 0.8, 1}
\pdfoutput=1

\begin{document}

\title[A review on locomotion robophysics]
{A review on locomotion robophysics: the study of movement at the intersection of robotics, soft matter and dynamical systems}

\author{Jeffrey Aguilar, Tingnan Zhang, Feifei Qian, Mark Kingsbury, Benjamin McInroe, Nicole Mazouchova, Chen Li, Ryan Maladen, Chaohui Gong, Matt Travers, Ross L. Hatton, Howie Choset, Paul B. Umbanhowar, Daniel I. Goldman}

\address{School of Physics, Georgia Institute of Technology, Atlanta, GA, USA}
\ead{daniel.goldman@physics.gatech.edu}
\begin{abstract}
{\bf SHORT ABSTRACT}
In this review we argue for the creation of a physics of moving systems -- a locomotion ``robophysics'' -- which we define as the pursuit of the discovery of principles of self generated motion. Robophysics can provide an important intellectual complement to the discipline of robotics, largely the domain of researchers from engineering and computer science. The essential idea is that we must complement study of complex robots in complex situations with systematic study of simplified robophysical devices in controlled laboratory settings and simplified theoretical models. We must thus use the methods of physics to examine successful and failed locomotion in simplified (abstracted) devices using parameter space exploration, systematic control, and techniques from dynamical systems. Using examples from our and other's research, we will discuss how such robophysical studies have begun to aid engineers in the creation of devices that begin to achieve life-like locomotor abilities on and within complex environments, have inspired interesting physics questions in low dimensional dynamical systems, geometric mechanics and soft matter physics, and have been useful to develop models for biological locomotion in complex terrain. The rapidly decreasing cost of constructing sophisticated robot models with easy access to significant computational power bodes well for scientists and engineers to engage in a discipline which can readily integrate experiment, theory and computation.

\newpage
{\bf EXTENDED ABSTRACT}
Discovery of fundamental principles which govern and limit effective locomotion (self-propulsion) is of intellectual interest and practical importance. Human technology has created robotic moving systems that excel in locomoting on and within environments of societal interest: paved roads, open air and water. However, such devices cannot yet robustly and efficiently navigate (as animals do) the enormous diversity of natural environments which might be of future interest for autonomous robots; examples include vertical surfaces like trees and cliffs, heterogeneous ground like desert rubble and brush, turbulent flows found near seashores, and deformable/flowable substrates like sand, mud and soil. Here we argue for creation of a physics of moving systems -- a locomotion ``robophysics'' -- which we define as the pursuit of the discovery of principles of self generated motion. Robophysics can provide an important intellectual complement to the discipline of robotics, largely the domain of researchers from engineering and computer science. The essential idea is that we must complement study of complex robots (and animals) in complex situations with systematic study of simplified robophysical devices in controlled laboratory settings and simplified theoretical models. We must thus use the methods of physics to examine successful and failed locomotion in simplified (abstracted) devices using parameter space exploration, systematic control, and techniques from dynamical systems. Using examples from our and other's research, we will discuss how such robophysics studies have begun to aid engineers in the creation of devices that begin to achieve life-like locomotor abilities on and within complex environments, have inspired interesting physics questions in low dimensional dynamical systems and soft matter, and have been useful to develop physical models for biological locomotion. Examples include hexapedal locomotion on hard ground, aerial locomotion of fly-size robots, aquatic propulsion and station-keeping by ribbon fins, and undulatory subsurface sand-swimming. Our studies also reveal surprising benefits of studying locomotion on granular media in experiment and computation/theory, despite a limited understanding of interaction mechanisms in these substrates.  From a theoretical point of view, we will discuss how the field of geometric mechanics can provide a framework and a language to understand the character and principles of effective locomotion. The rapidly decreasing cost of constructing sophisticated robot models with easy access to significant computational power bodes well for scientists and engineers to engage in a discipline which can readily integrate experiment, theory and computation; such convergences bodes well for prediction, design and understanding of success and failure of self-propelling automata in complex environments. 


\end{abstract}

\maketitle

\section*{Introduction}

\begin{shaded}

``
...I will, however, maintain that we can learn at least two things from the history of science. One of these is that many of the most general and powerful discoveries of science have arisen, not through the study of phenomena as they occur in nature, but, rather, through the study of phenomena in man-made devices, in products of technology, if you will. This is because the phenomena in man's machines are simplified and ordered in comparison with those occurring naturally, and it is these simplified phenomena that man understands most easily... Thus, the existence of the steam engine, in which phenomena involving heat, pressure, vaporization, and condensation occur in a simple and orderly fashion, gave tremendous impetus to the very powerful and general science of thermodynamics. We see this especially in the work of Carnot. Our knowledge of aerodynamics and hydrodynamics exists chiefly because airplanes and ships exist, not because of the existence of birds and fishes. Our knowledge of electricity came mainly not from the study of lightning, but from the study of man's artifacts.''  John Pierce~\cite{pierce2012introduction}.

\bigskip

``What I cannot create, I do not understand'', Richard Feynman

\end{shaded}
Nearly 100 years have passed since the word ``robot'' was coined by Karel Capek in his play R.U.R.\ (Rossum's Universal Robots)~\cite{vcapek2004rur} (derived from the Slavic word, ``robota'', synonymous with servitude, forced labor and drudgery~\cite{scifri}). In that time, largely due to the efforts of engineers and computer scientists, our society has indeed become permeated with a certain kinds (largely virtual or immobile) of robots: from ATMs to factory welding robots to smart thermostats to ``assistants'' like Siri. The dream of creating robots that can interact gracefully in the physical world with humans remains science fiction; but due to a convergence of low cost actuators, sensors, computing resources, and storage~\cite{markoffbook}, increasingly complex robots are soon to enter our lives in a more physical way, as the devices we wear, which assist us when we fall, drive us on roads and other terrain, or help us change light bulbs. A key ingredient enabling robots to work with us and for us will be robust mobility in all environments, so they can assist not as only rolling vacuum cleaners on hard floors, but as drainpipe inspectors, search and rescue agents in rubble, mountain, desert environments, beach lifeguards and package deliverers.

In our era, such robust all-terrain self-propelled movement remains the province of biological living systems: from cells to mice to elephants, organisms negotiate terrain no engineered device can. In recent years, human-scale robots have even begun to develop capabilities resembling those of animals (Fig. 1)  but performance in natural environments (trees, rubble, gusty environments) is typically low relative to biological systems. Given our ability to make cars and jet airplanes move well (in certain environments, far better than animals!), this contrast in abilities seems odd. However, our best mobile engineered devices confront relatively simple environments (like paved roads), allowing for relatively simple mechanical systems and controls.  In contrast, animals encounter a huge diversity of environments, and have evolved morphology, controls and materials to enable robust navigation in air, water, mucous, dirt and sand. Animals (as well as robots in certain environments~\cite{bdcheetah,ramezani2014}) often locomote so beautifully (and seemingly efficiently) we can be lulled into a false sense of security that principles and mechanisms are understood (or if not understood, trivial or unimportant to discover!).

While largely unexplored in a systematic way by the robotics engineering and computer science communities, studies seeking the underlying principles of effective locomotion in natural environments have flourished under scientific inquiry, particularly among biologists and mathematicians. Such efforts have led to a number of developments in characterization and modeling of locomotion which have been the focus of books and reviews ~\cite{alexanderbook,bottcher2006principles,sfakiotakis1999review,dicAfar,pfeifer2006body,Pfeifer2007,holAful06,Ijspeert2014,vogel2013comparative,goldman2014colloquium,hosoi2015beneath}. Of particular note is the efforts of biologists and mathematicians (with some physicists) in the development of the study of ``biofluids'' locomotion problems~\cite{lighthill1975mathematical,childress1981mechanics,dudley2002,chiAdud,Alben2005,childress2012natural,vogel1996life}, which typically have uncovered fascinating aspects of locomotion in fluids and physics relevant to fluid dynamics, but with less of a focus on general principles of control of movement. Such focus tends to be found in the biological/physiological literature which develop complex theoretical models, often with tens to hundreds of parameters. While such models can give insight~\cite{panAzaj,mcmillen2008nonlinear,tytell2010interactions,bobbert2001dependence,ijspeert2001connectionist,ghigliazza2004minimal,seipel2004dynamics,seipel2007simple,holAful06} they make it challenging to elucidate fundamental principles. 

Relative to the effort expended by biologists (including physiologists) and mathematicians, the study of movement in complex terrain has received less attention from physicists. We suspect that to the intellectual descendants of Galileo and Newton, problems associated with a classical body moving from A to B either seem simple (wheel rolling), solved (Newtonian mechanics) or far too complicated (sand-swimming). From our (a group which has been studying locomotion of animals and robots on complex terrain for over a decade) point of view, the current grasp on the principles of locomotion in modern robotics (and biology) is reminiscent of Carnot's thoughts on the understanding of steam-powered engineering at the advent of thermodynamics: {\em ``Notwithstanding the work of all kinds done by steam-engines, notwithstanding the satisfactory condition to which they have been brought to-day, their theory is very little understood, and the attempts to improve them are still directed almost by chance.''}~\cite{carnot1890reflections}. In fact, the works of Carnot~\cite{carnot1890reflections}, Kelvin~\cite{kelvin1851dynamical}, Clausius~\cite{clausius1850bewegende} were inspired by a desire to understand the specific phenomena governing the technological improvements of engines and machines which turn steam power into motion. In turn, it required the creation of a new fundamental theory of nature, the science of  thermodynamics, to unify the understanding of these phenomena. These words are inspiring to a physicist seeking general principles and perhaps new emergent phenomena associated with movement.

In this review, we will argue that building on the foundation established by biologists, mathematicians and engineers, the application of the methods and philosophy of physics can add at least two developments to help create a fundamental understanding of the necessary and sufficient conditions (a ``minimal feature set'') for robust self-propelled locomotion in the real world. The first development echoes the thoughts of Pierce (above quote) and relies on relatively recent efforts to understand the richness of biological movement through the controlled and systematic study of man-made robotic locomotors. Over the past 25 years, various groups have developed robots {\em physical models} of organisms and used the models to discover principles by which organisms move through air and water, on land and, in our area of focus, through granular media. Such physical models are useful scientifically because they allow systematic studies on features of morphology and control in ways that are often impossible with living systems. These physical models can also form the basis for the creation and understanding of more complex real-world robots, and can more broadly demonstrate fascinating new emergent phenomena in classical dynamical systems. The second development, which we will argue provides the elements of a dynamical systems language of locomotion (i.e., fundamental principles of movement), arose in the late 1980's when Shapere and Wilczek~\cite{shapere1989geometry, shapere1987self} recognized that certain types of movement, namely those involving shape changes in highly dissipative media, could be described by a ``gauge potential'' framework. This framework has already proved useful as a language for locomotion, largely due to engineers who embraced and developed it~\cite{marsdenreviewgeo,ostrowski_98,Kelly:1995,boyergeometric2015} to become predictive in relatively simple environments~\cite{Hatton:2011IJRR,bullo2001kinematic} and recently in a more complex medium~\cite{hatton2013geometric}.

\begin{figure}[h]
\begin{centering}
\includegraphics[width={1\hsize}]{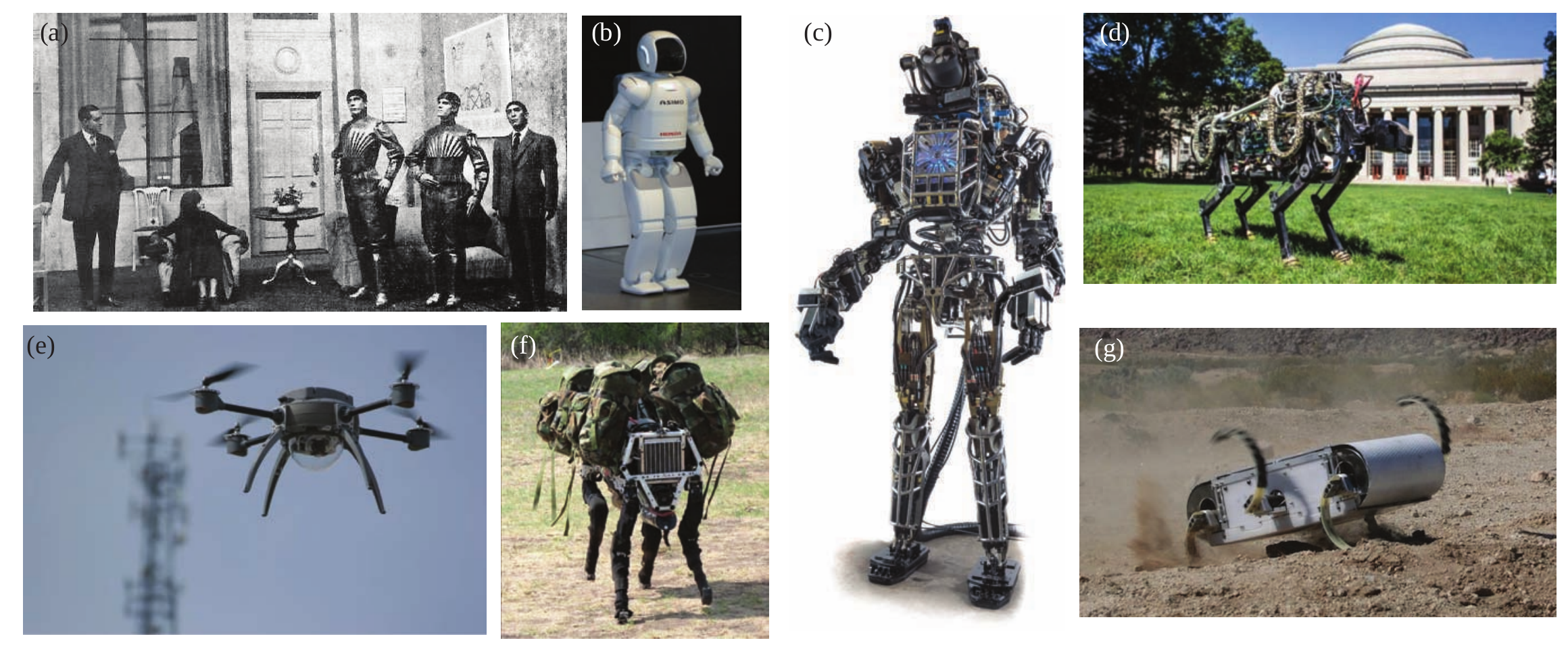}
\caption{\label{fig:robots} Robots have entered our lives and environments. (a) A scene from R.U.R. (Rossum's Universal Robots)~\cite{vcapek2004rur}. Recent engineering advances have made strides in realizing the vision of mobile robots that are integrated in the complex environments of our society, from (b) Asimo and (c) Atlas to (e) delivery and video capture drones, to (d) Cheetah as well as (f) BigDog and (g) RHex robots, both capable of traversing relatively simple external environments. Research and development of drones, RHex and BigDog have expanded their mobility to a variety of increasingly (f,g) complex terrains.}
\end{centering}
\end{figure}

\section{Foundations of locomotion robophysics: key ingredients in a physics of moving systems}
\label{sec:found}
We propose that the above developments form the beginnings of (and demonstrate the need for) a physics of moving systems -- a ``locomotion robophysics''. The latter term we define as the pursuit of fundamental principles governing movement and control of self-deforming entities (presently electromechanical, hydraulic or pneumatic, but in the future perhaps made of pressurized elastomer networks~\cite{wehner2014pneumatic}, carbon nanotubes~\cite{aliev2009giant} or amplified piezo arrays~\cite{schultz2010intersample}) interacting with complex environments. These are fundamentally non-equilibrium processes where motion emerges through effective interaction via internally driven actuation as opposed to equilibrium systems which naturally find free energy minima through externally driven forces. We (as a group of experimentalists) feel that robophysics builds on a discipline which has informally been referred to as ``experimental robotics''~\cite{buehlerprivate,experimentalroboticsbook}, and will likely be crucial to the discovery of novel locomotor principles.  To be specific, we now present a list of what we view as important aspects of robophysical study, our proposed {\em foundations of robophysics}:

\subsection{Integration of science and engineering: creation leads to understanding and vice versa}
As noted by Pierce in the introductory quote, the history of two-way interaction between physics and engineering is quite rich, and both disciplines have benefited: examples include fluid dynamics and the design and understanding of airplanes and ships, thermodynamics/statistical mechanics and improvements in engines, understanding genetic circuits~\cite{hasty2002engineered} and even quantum mechanics and improvements in ovens and lightbulbs~\cite{kragh2002quantum,heilbron1987dilemmas}. Thus we propose that the intersection of engineering, physics and biology will continue this mutualistic interaction (and in successful collaborations, mutualistic teaming ~\cite{full2015interdisciplinary}) in pursuit of understanding locomoting systems. Echoing (and perhaps distorting) Feynman's words, we argue that creating real-world engineered devices is critical if we are to claim that we understand the robophysical phenomena observed in the laboratory; yet conversely, simple robophysical models are also critical to truly understand the phenomena that real-world engineered robots encounter in natural environments. To fully understand the physics principles discovered in the lab with experimental robophysical devices, it is important to test those principles with real world robots in natural environments that approximate laboratory conditions. Observing how these principles play out in more complex devices (or animals) can lead to observations of novel phenomena which are not observed in more controlled laboratory settings; such phenomena then inspire new robophysics work.

\subsection{Simplification aids understanding}
To facilitate the discovery of locomotor principles, we argue that the methods of experimental physics, e.g., joining theory with systematically controlled laboratory experiments on simplified systems, are essential. Studying complex engineered devices and/or attempting to model every detail of the natural world often results in intractable complexities. Thus, the locomotion robophysics approach is to experiment on simplified (typically in the laboratory) devices which contain the minimal geometric and control design necessary to produce dynamics similar to the complex locomotors (robots or animals) from which they are derived; the robophysical devices thus function as models of the more complex locomotors. In contrast to bio-mimetic~\cite{kato2007bio,paulson2004biomimetic} robots which often replicate the appearance of the entire organism, robophysical devices function as models and are thus often abstracted from specific features of appendage-induced locomotion (e.g., leg, flipper\cite{mazouchova2013flipper}, tail~\cite{Libby2012,johnson2012tail,kohut2013precise,kohut2013aerodynamic}, fin~\cite{Esposito01012012}, etc.). 

In robotics and biology, this approach has been most fruitful in the study of legged hopping robots, in which analysis of the simplified Spring Loaded Inverted Pendulum (SLIP) model~\cite{holAful06} and associated robots \cite{Raibert.Book1986,Akinfiev2003} has facilitated the study of hopping, jumping and running gaits. More broadly, robophysical models have been useful for comparative studies~\cite{Ferrell1995,Schroer2004} and systematically testing hypotheses of biological locomotion. An example from our work illustrates this approach. In a study of locomotion of a hexapedal robot on granular media, Li et al~\cite{li2013terradynamics} chose to study a vastly simplified and smaller model of the RHex robot, which did not have the complex controller or limbs present in the field-capable device. However, this robot proved amenable to systematic control (see next section) of certain parameters and enabled development of a resistive force theory which has now been successfully applied to the larger more realistic robot~\cite{Qian2015BIBM}. As another non-biologically motivated example, in the study of wheeled locomotion, replacing complex vehicle-terrain interaction with a single wheel allowed for a better understanding of locomotor terramechanics in wheeled vehicles~\cite{Ding2013}. 

\subsection{Embracing poor performance and systematically surveying parameter space}

Engineering has a long tradition in optimization and failure avoidance. In particular, failure analysis has been an important branch of engineering in designing products and materials that can avoid failures during regular use through the application of systematic methods that explicitly investigate and understand the causes of such failure modes~\cite{brooks2002failure,anderson2005fracture,martin1999electronic}. In locomotion robotics research, a common approach is to engineer robots with a focus on successful task completion and then refine such ``good'' performance through a process of optimization. Machine learning techniques (see these papers~\cite{atarilearning,schulman2015trust} for exciting recent examples) have been effective to this end, often times focusing on the challenge of avoiding local minima or maxima of optimality which may not necessarily lead to global robustness. This has worked quite well: engineers already create devices with whimsical names like BigDog~\cite{plaAbue06}, RHex~\cite{saranli2001rhex} (Fig.~\ref{fig:robots}f,g) and Atlas~\cite{kuindersma2015optimization}(Fig.~\ref{fig:robots}c, also see Petman~\cite{GabeNelson201230_372}) that can locomote on relatively complex terrain despite the lack of understanding of the dynamics of the robot interacting with the environment. Yet, while a focus on optimization and safety factors can clearly help mitigate failures and ensure success in specific situations, neglecting a study of fundamental locomotor principles comes at a certain cost. Feedback control schemes based on increasingly sophisticated on-board sensing and computing platforms~\cite{astrom2010feedback} can be optimized specifically with disturbance rejection of environmental unknowns in mind, yet it is often impossible to know if such robots will always perform well in a diversity of environments (and why they fail when they do) without an understanding of the dynamical system that is comprised of the physical device, its controller and interactions with the environment.

Robophysics, on the other hand, does not distinguish failure from success, as the goal is not necessarily optimization or the avoidance of failure, but rather the principles of environment-locomotor interaction that produce different performance outcomes. Studying failure modes and success using the methods of robophysics can reveal the subtlety and beauty in how non-failing behaviors emerge. In addition, a broad exploration of a parameter space can lead to a greater understanding of the mechanisms of optimal performance that can then be extended to a wider range of scenarios (through nondimensionalization, for example). A few examples from our own robophysical work serve as examples by which this process can produce new mobility: while the RHex robot displayed beautiful ability to bounce across hard ground and some ability to walk across granular substrates, the reason different parameters in the limb controller worked well were unknown. It was not until we undertook systematic studies of performance on granular media of varying compaction that we revealed the robophysical mechanics underpinning RHex's locomotor abilities, and in addition, why the locomotion was so sensitive to aspects of the ground and the controller~\cite{li2009sensitive}.
As another example,  a robotic snake~\cite{wright2007design} effective in many environments including hard ground, pipes, poles and water was unable to climb sandy slopes without catastrophic slips until systematic experiments revealed that contact length of the snake must be properly modulated to prevent yielding, maintain balance, and reduce stress on the granular substrate, ~\cite{marvi2014sidewinding} (Section~\ref{sec:sidew}). Since then, we have been able to create designs and new control schemes that improve robustness of these devices. And in a study on robotic sand jumping~\cite{aguilar2015jumpsand}, various jumping strategies were surveyed, one of which performed poorly; further analysis revealed the mechanism for failure was an added mass effect that had never been confirmed in dry granular media, which has implications about fast locomotion on granular media beyond jumping. 

\begin{figure}[ht]
\begin{centering}
\includegraphics[width={.67\hsize}]{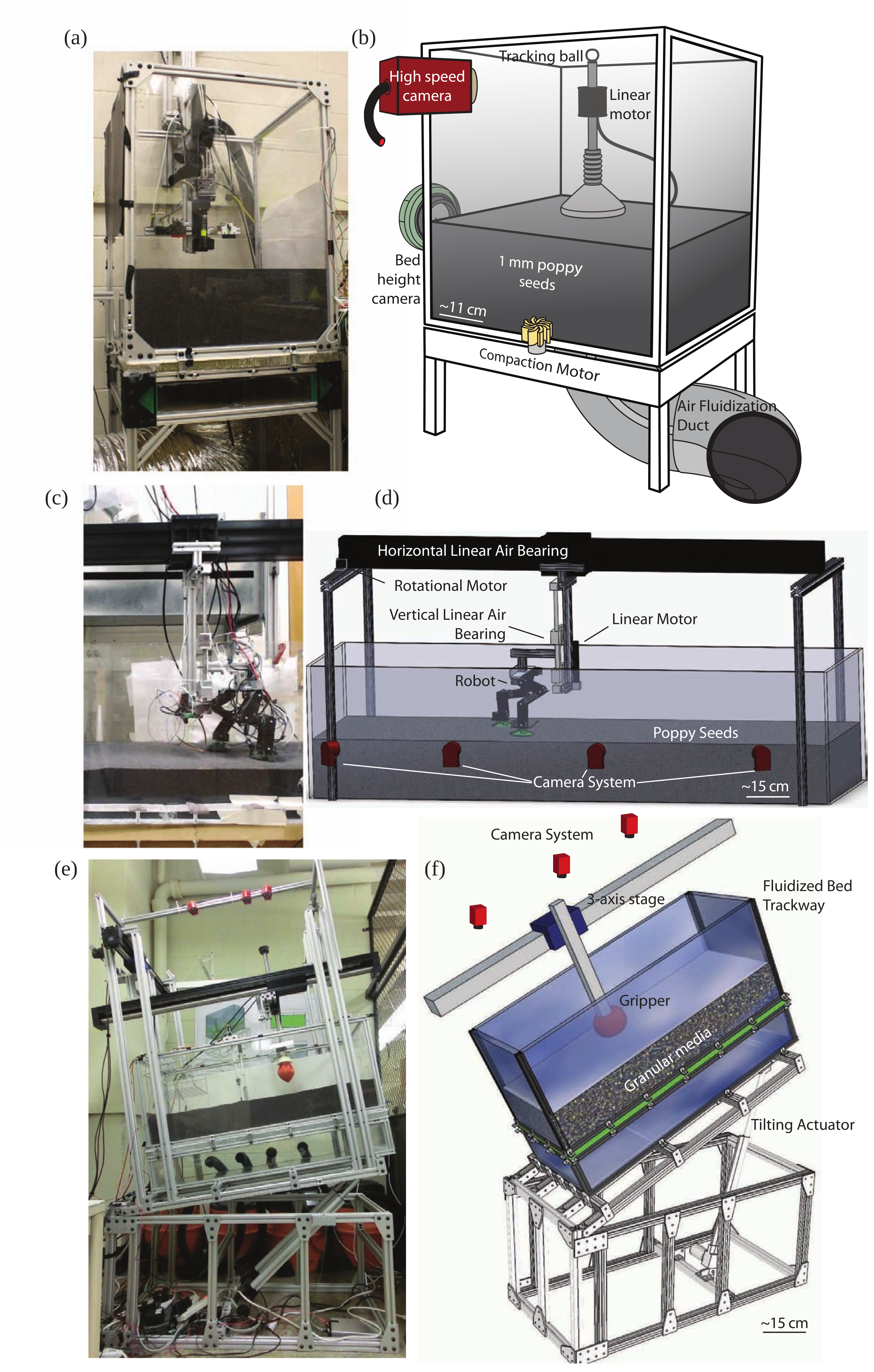}
\caption{\label{fig:auto} Automated robophysics experiments from our group. (a,b) A simple jumping robot is constrained by an air bearing to jump vertically in the center of a bed of poppy seeds with a volume fraction controlled by a blower that forces air through the bottom. Exploration of paramater space has revealed rich nonlinear dynamics on hard ground~\cite{aguilar2012lift} (Section~\ref{sec:simple}) and, by iterating through various jumping strategies on different granular states, is being used to uncover new granular physics (Section~\ref{sec:highspeed}) and its effect on fast locomotion.  (c,d) An apparatus that cycles through different bipedal walking gaits on granular media; the robot is constrained by low friction bearings to walk in the vertical plane inside a fluidized bed. (e,f) The Systematic Creation of Arbitrary Terrain and Testing of Exploratory Robots (SCATTER) system allows for automated three axis manipulation of heterogeneities such as boulders as well as robots traversing a granular substrate with specified ground stiffness and surface incline.}
\end{centering}
\end{figure}

Systematically surveying a parameter space -- while critical to compare theory and experiment, discover transitions and bifurcations in dynamics, and understand transitions in performance modes -- has an inherent repetitiveness which leads to a labor-intensive endeavor when manually executed by humans. Taking inspiration from the methodologies of physics (largely from fields like nonlinear dynamics and condensed matter physics), our group has begun a program of automating our experiments, in effect, creating robots that control robots and their locomotor environments. We will illustrate how such automation is becoming a critical feature of robophysics, allowing for high throughput parameter variation. A few examples from our work are shown in Figure~\ref{fig:auto}. Such automation then allows systematically exploring parameters and identifying the most important parameters of more complex systems~\cite{machta2013parameter}.

\subsection{The importance of feedback and control templates}
\label{sec:control}

The earliest widespread use of feedback control (or at least its onset of popularity and interest among engineers) can be traced back to the 18th century, with the centrifugal governor invented by James Watt (first analyzed by Maxwell~\cite{maxwell1867governors}) to regulate the speed of the steam engine, where the intelligence of the controller was embedded in the passive mechanics of the device~\cite{mayr1970origins}. These early devices were later treated to in-depth analyses of the physics principals on which they operate~\cite{airy1840regulator}. Today, much effort in the locomotion robotics community has been spent in developing feedback control strategies to generate stable movement patterns in different environments. However, from a robophysics perspective, these approaches can leave unexplored important questions regarding high-level (behavioral) objectives of control and learning, many of which were anticipated by the mathematicians von Neumann and Wiener (and others) ~\cite{heims1980john} during the heyday of the cyberneticians~\cite{wienerbook} (with `cyber' adapted from {\em kybernan}, the Greek word meaning to steer or govern~\cite{cybernet}). We argue that utilizing the methods of robophysics in conjunction with a particular approach to locomotor control can advance the control of movement in the real world.

Our approach to locomotor control of movement is encapsulated in the ideas developed by Full and Koditschek in their seminal paper~\cite{full1999templates} and nicely discussed in a control theory framework in~\cite{roth2014comparative}. In this work, to tame the complexity of understanding the control of movement in complex animals, the authors proposed to formulate the modeling process into a hierarchical structure of complexity, consisting of ``templates'' and ``anchors'', where templates are reduced order models with a minimal number of parameters that encode the behavior of more complex locomotors and prescribe high-level control strategies to achieve such locomotor behavior~\cite{LABid2660138,schAhol,LABid2661112,seyfarth2000optimum,alexander1995leg}. Anchors are the more detailed and complex underlying mechanistic systems which can be controlled to follow the template. Adding the robophysics approach to the templates/anchors scheme can be an invaluable tool to create life-like locomoting robots (which are increasingly complex with emergent dynamics that defy a reductionist understanding). Such an approach is very much aligned with the spirit of Pierce's quote in our introduction which seeks to study simplified models (here of animals and robots) to discover fundamental principles. For the physicist, addressing questions in the control of movement from a template/anchor plus robophysical approach will generate interesting (and likely novel) dynamical systems that incorporate not only Newton's laws applied to the robot and environment, but also computational laws -- the models that ultimately define and limit the feedback schemes. We will argue that recent developments in the field of geometric mechanics (see Section \ref{sec:geomech}) hint at a general framework to discover templates, e.g. ~\cite{HattonClawar:2012,PhysRevLett.110.078101}. However, understanding closed-loop locomotor efficacy in complex environments will require more than the combination of control theory, dynamics and soft matter physics; as the reader has no-doubt surmised, our group argues that systematic experiments are critical!




Our work in sidewinding locomotion exemplifies of the power of this approach. As we will discuss in more detail in Section \ref{sec:sidew}, systematic experimentation with a snake robot (as well as biological observations) revealed that sidewinding locomotion, complicated though it may seem, can be generated by a relatively simple template consisting of superimposed vertical and horizontal propagating waves of body curvature at $\pi/2$ phase offset from each other~\cite{Henry_PNAS,marvi2014sidewinding}. This two-wave model was originally developed by the engineers in our group to effect locomotion of their robot on hard ground. However, once we identified this pattern as a template (which is also followed by the robot's biological counterpart, the sidewinder rattlesnake), it became clear how a sand-dwelling limbless locomotor benefits from controlling its shape to follow this pattern. For example, we discovered that sandy slopes could be effectively climbed by modulating the vertical wave to elicit a solid-like response from the sand~\cite{marvi2014sidewinding}. Subsequent studies revealed how other modulations (amplitude and phase) of the two-wave template generated sidewinding turning behaviors~\cite{Henry_PNAS}. Using modulations of the template to drive behavior greatly simplifies locomotion in this system (see the example of maze-following in ~\cite{Henry_PNAS}). And with the template (and its modulations) in hand, we can understand the necessary ingredients (motors, sensors, skin) the anchor must possess to generate these diverse behaviors (and thereby close the loop to develop autonomous robots). For example, during movement on inclined granular media, encoders for each actuator may be sufficient to control the body to follow a template, but must be augmented with sensors that can determine body slip and average inclination.

\section{Robophysics as an enabler for progress in diverse areas}
\label{sec:diverse}

The study (and use) of engineered robotic systems will not only enable progress in a diversity of physics disciplines such as soft matter and complex non-equilibrium systems (as well as biological locomotion), but such progress will also be rapidly accelerated through automation granted by robotics. Conversely, the methodologies native to physics will help illuminate general principles of robot locomotion relevant to robot control and design (and biological systems). Once the principles are worked out in the robophysical system, they can be instantiated and tested in more robust devices or sought out in organisms. We now list a few areas that we predict will benefit from the robophysics discipline.

\bigskip

\noindent {\em Dynamical and nonequilibrium systems}: The locomotion of robots through an environment can serve as an excellent model system to advance our understanding of the features of complex dynamical systems driven far from equilibrium (like robustness and complexity~\cite{Carlson19022002,Goldenfeld02041999} as well as emergenent phenomena~\cite{anderson1972more}). More down to earth, the study of robots as dynamical systems is a sure way to create novel inquiry into dynamical systems, particularly when aspects of feedback control are considered ~\cite{astrom2010feedback}, as has been demonstrated in the robotics and applied mathematics communities (e.g. periodic orbits in robot hopping~\cite{kodAbue91} and juggling~\cite{Buehler01041994,rizzi1992progress}, stability of hybrid dynamical systems in running robots~\cite{holmes2006dynamics}). In fact, themes and tools from literature studying low and high dimensional dynamical systems (phase portraits, bifurcations, limit cycles, chaos, pattern formation, normal forms) form a useful language and conceptual framework/underpinnings for understanding stability and control of locomoting systems. And experimental inquiry into robophysical systems can certainly help to test theoretical predictions and challenge and introduce new concepts. For example, we will emphasize recent (since the 1980s) advances in geometric mechanics, originally proposed as a framework to study locomotion at low Reynolds number, but since used to frame a wide class of problems including self-righting cats, reorienting satellites, the snakeboard\cite{ostrowski1998geometric}, and effective locomotion of sand-swimming robots through proper ``self-deformation'' patterns (see Section~\ref{sec:geomech}). From our point of view, experimental tests and inquiry have not kept pace with these exciting theoretical advances. And because they can be decomposed into a hierarchy of systems with increasing dimension, locomoting systems can also provide excellent tests of ``mesoscopic'' dynamical systems composed of a large number of degrees of freedom. For example, studying the locomotor stability of a robot composed of increasingly complicated actuation and control mechanisms poses questions into the sensitivity of gross behaviors to lower level parameters~\cite{kagami2002,chevallereau2009,grizzle2014, kuo1999}. Perhaps the work on ``sloppy'' models~\cite{machta2013parameter,gutenkunst2007universally} will be applicable here; locomoting robophysical systems can provide an arena to experimentally test such theories.

\bigskip

\noindent {\em Soft matter physics}: Locomotion robophysics promises to advance our understanding of terrain composed of materials that are neither rigid solids nor Newtonian fluids (like sand, snow and grass) with complex rheological response to interaction, that is ``soft matter'' systems. As advocated in~\cite{krotkov1990active}, in which ``every step is an experiment'', robophysics can aid in the discovery of new soft matter physics by revealing the unique locomotor dynamics that arise from robot-substrate interactions, which can inspire systematic studies of these often novel interactions with soft matter~\cite{zhang2014effectiveness}. The equations of motion of most terrestrial substrates are not yet known, and thus {\em require} experimental methods to understand how movement emerges from interaction with these environments. Further, such materials and interactions are ``non-standard'' in the soft-matter literature yet ubiquitous in natural environments substrates, thus offering a rich array of systems to be studied. 

Once such (typically complex) interactions are identified, the machinery of soft matter physics can be applied to systematically develop empirical understanding of responses. One of our recent studies~\cite{aguilar2015jumpsand} (briefly discussed in Section \ref{sec:highspeed}) makes this abundantly clear: while impact into dry granular media has been studied in different regimes for hundreds of years, new physics associated with jamming and added mass were revealed only after analyzing the sand jumping dynamics of a 1-D legged robot. Unlike standard impact~\cite{Katsuragi2007} and constant speed intrusion~\cite{li2013terradynamics} experiments, jumping produced much more complex granular interactions that, when analyzed, revealed new transient dynamics which significantly influence locomotor performance. Other examples in dry granular media we will discuss include ``disturbed ground'' in the FlipperBot work~\cite{mazouchova2013flipper} (Section~\ref{sec:flip}) or resistive forces on sandy slopes like found in the sidewinding snake study~\cite{marvi2014sidewinding} (Section~\ref{sec:sidew}). And such empirical insights from experiment can inspire new theoretical advances (see~\cite{askari2015intrusion} for a recent example based on granular resistive force theory). A better understanding of interaction with soft matter will lead to a ``terradynamics''~\cite{li2013terradynamics} of land-based locomotion (in analogy with hydro- and aerodynamics which allow submarines and planes to explore water and air) and allow robots to robustly and independently explore disaster sites, comets and other celestial objects.


\bigskip

\noindent{\em Biological movement and principles of robustness in living systems}:  As noted throughout the introduction, a robophysics of locomotion can advance our understanding of biological movement. Robophysics can thus be a critical element in understanding aspects of movement within a more broad ``physics of living systems''. Such advances can result from physicists' tendency toward instrumentation and model building, but also from the desire in the discipline to focus on simplification as a tool to understand complex systems (e.g. finding the ``hydrogen atom'' of the problem).  Because of this, robophysics can provide a practical starting point to discover important functional principles~\cite{bialek2012biophysics} in living systems (building on, of course, foundational work in cybernetic~\cite{walter1950mitation} and physiological modeling as well as the study of organism movement using the tools of dynamical systems ~\cite{holAful06,LABid2660138,schAhol,LABid2661112,seyfarth2000optimum,alexander1995leg}). Locomotion robophysics can be used to confront issues of biological robustness~\cite{Carlson19022002,barkai1997robustness,hartwell1999molecular,kitano2004biological} whereby complex organisms are studied in terms of the hierarchical and heterarchical organization of their mechanical and control structures.  Locomotion robophysics can also provide ``hands-on'' systems to examine modeling approaches in systems composed of many degrees-of-freedom, common in many areas of science and engineering (e.g. see for example~\cite{gutenkunst2007universally} for a discussion of complex models in systems biology).

Specifically, building on advances in physical modeling of organisms, robots (and experimentally validated simulations) can be used to test biological hypotheses concerning body and limb morphology, neuromechanical control hypotheses~\cite{ding2012mechanics,nishikawa07}, including the systematic search for templates and anchors~\cite{full1999templates} in physiologically relevant environments. These systems are much simpler and amenable to analysis, and require fundamental study of dynamical systems, complex networks, among other topics.  The robots function as physical models, which as argued above are, when models of the environment are unavailable, are perhaps the {\em only} methods to model organism movement.  In this way, robophysics in biology subscribes to Feynman's philosophy highlighted in our introduction --  biological systems (nor complex engineered devices) cannot be {\em understood} without building or creating models whose goal is to reveal fundamental, cross-cutting principles by which living systems move with such efficacy in a staggering array of natural substrates. 

The modeling framework of templates and anchors~\cite{full1999templates} provides a powerful framework by which robophysics can aid biology from the ``top-down'' through studying principles of life using robots, and perhaps can be on occasion more tractable than by dismantling biological systems (synthesis vs. reduction).  We argue that directly probing ensemble processes in complex biological structures proves too intractable to develop meaningful and predictive models of locomotion. In fact, a number of physicists are thinking about behaviors in a similar way to the approach in~\cite{full1999templates}. Members of Bialek's group have contributed significantly to identifying how these templates emerge in insects as stereotyped template behaviors~\cite{berman2014mapping,stephens2008dimensionality}. Using the methods of robophysics, we can aid biology in testing and understanding these templates and anchors by physically instantiating them with robots. Such robots, as physical models, are a much richer representation of living systems than mathematical models alone, however, both must be studied in parallel.  Such an approach is also able to systematically probe the effects of progressively increasing locomotor and environmental complexity (extra appendages, geometric features, environment heterogeneity, etc.).


 
\bigskip

\noindent{\em Improved real-world robots}: And finally, robophysics will help discover principles why our best robots significantly under-perform living locomoting systems in arbitrary terrain; and how to improve the performance of engineered devices in real-world environments. Thus from the engineering perspective, the robophysics approach can also help us rethink how we approach robotics: we can begin to break the seemingly endless cycle building and programming new iterations of field-ready robots that barely outperform their predecessors. Already (as we will show) our robophysics studies demonstrate  that we do not have to engineer hardened robots to do beautiful robot experiments.  Perhaps focusing on simplified model systems can be a more practical and time (and money) saving way to unlock truly advanced mobility.  Once the principles are studied in a laboratory setting and once important physics of dynamical systems and soft matter and controls are learned, these lessons can then be used to guide the design of hardened devices that are actually capable of being deployed to search, rescue, and explore. 

\bigskip

\section*{Specific examples to be addressed in this review}

The remainder of our review is organized around specific examples involving single robots interacting with environments that create challenges for scientists and engineers and which encapsulate the foundations of locomotion robophysics as delineated above. Our examples are of course not exhaustive (and we hope will not prove exhausting) but serve to illustrate the effectiveness of the robophysics approach in different environments, whereby the use of simple robot models facilitate the iterative comparison between experiment, simulation and analytical theory. Our review will focus on terrestrial locomotion robophysics, largely because we feel this is a neglected area and one in which our group has made recent progress in the robophysics vein.  We will predominantly focus on steady rhythmic interactions with the environment, a tiny subset of possible behaviors (impulsive jumps, jerks, etc.). In addition to sections describing progress we have made in specific environments, we will also highlight (in blue ``sidebar'' sections) other fields which benefit from robophysical studies, such as paleorobotics and biomechanics. We will also comment on computer simulation as a critical component which, when validated against experiment, allows for computer models of movement, and access to quantities that are challenging to measure in experiment. And we will devote a section to geometric mechanics, as we feel this framework is appealing to physicists as it allows understanding the character of locomotion. Following our introduction above, we will discuss:

\begin{itemize}
  \item \textbf{Section \ref{sec:simple}: Locomotion robophysics in hard ground environments:}
We discuss various studies of robots hopping, running and jumping on hard ground; such studies have led to arguably the most successful legged robots in operation today. In particular we highlight how one of our first experiments to be fully automated reveals that complex dynamics emerge from even the simplest robotic system, a monopod hopper~\cite{aguilar2012lift}. In a later section (\ref{sec:highspeed}), we show how even this simple system challenges our current understanding of soft materials when we examine jumping on granular media.
  \item \textbf{Section \ref{sec:fluid}: Robophysics in air \& water:}
   We briefly review the use of robotics and simulation in understanding the physics of swimming and flying in Newtonian fluids.
  \item \textbf{Section \ref{sec:gm_robo}: Robophysics in granular media, movement on and within a substrate between solid and fluid:}
 We discuss our work on locomotion through dry granular media. In particular, we discuss how the frictional aspects of granular media dominate the movement of relatively slow robots, including a larger hexapod~\cite{li2009sensitive} (Section \ref{sec:rhex}), a robotic sidewinder~\cite{marvi2014sidewinding} (Section \ref{sec:sidew}), and a turtle-inspired device~\cite{mazouchova2013flipper} (Section \ref{sec:flip}); the disturbance and ``memory" of the ground plays a major role. We then discuss how an undulating robot, modeled after a sand-swimming lizard~\cite{Maladen2009}, can ``swim" in a self-generated and localized granular frictional fluid~\cite{maladen2011} (Section \ref{sec:sandswim}).
  \item \textbf{Section \ref{sec:hetsub}: Heterogeneous terrestrial substrates:}
  We discuss robophysics for locomotion in more complex granular substrates such as those with particles varying widely in size (e.g., powders to boulders). Substrates with such heterogeneities are enormously diverse; we will focus on the robophysics of single boulder ``scattering'' and maneuverability through simplified ``boulder lattices''.
  \item \textbf{Section \ref{sec:wet}: Wet terrestrial substrates:}
We briefly discuss robot locomotion through granular substrates which contain fluids; such materials are found in an enormous range of natural environments. And while little is known about locomotion in these materials, the physics and biolocomotion of and in such substrates is fascinating~\cite{mitarai2006wet,dorgan2007bmm,sharpe2015controlled}; we argue that studies like those pioneered in~\cite{winter2014razor} combined with improved experimental tools and theoretical understanding will advance this very important regime of locomotion robophysics. 
  \item \textbf{Section~\ref{sec:compu}: Computational tools in locomotion robophysics:}
  We discuss computational tools that have helped further explain and provide insight to robophysics experiments. In particular, we discuss the tools developed for fluids (\ref{sec:compufluid}), hard ground (\ref{sec:compuhard}), and granular media (\ref{sec:compugmslow} and \ref{sec:highspeed}). We discuss advances in granular resistive force theory (RFT) (Section \ref{sec:compugmslow}) which represents a new development in the continuum description of granular flows in the fluid/solid regime~\cite{Zhang2014}, as well as how the computational discrete element method (DEM) (Section \ref{sec:highspeed}) has been crucial to model aspects of robot locomotion in dry granular media. 
  \item \textbf{Section \ref{sec:geomech}: Geometric mechanics: a language of motion building blocks for robophysics:}
  We review how the mathematical geometric mechanics framework~\cite{marsdenreviewgeo,shapere1989geometry} can give insight into how arbitrary body translation and rotations result from cyclic self-deformations, even in substrates like dry granular media; we propose that geometric mechanics can serve as a general framework/language for locomotion robophysics.
   \item \textbf{Section \ref{sec:conclusion}: Conclusions and long term:}
   We conclude with discussion of future directions of robophysics, from the examination of locomotion through changing substrates in complex environments, to the consideration of feedback control models as an integrated component of the locomotor dynamical system.
\end{itemize}

\section{Locomotion robophysics in hard ground environments}
\label{sec:simple}

\begin{figure}[h]
\begin{centering}
\includegraphics[width={1\hsize}]{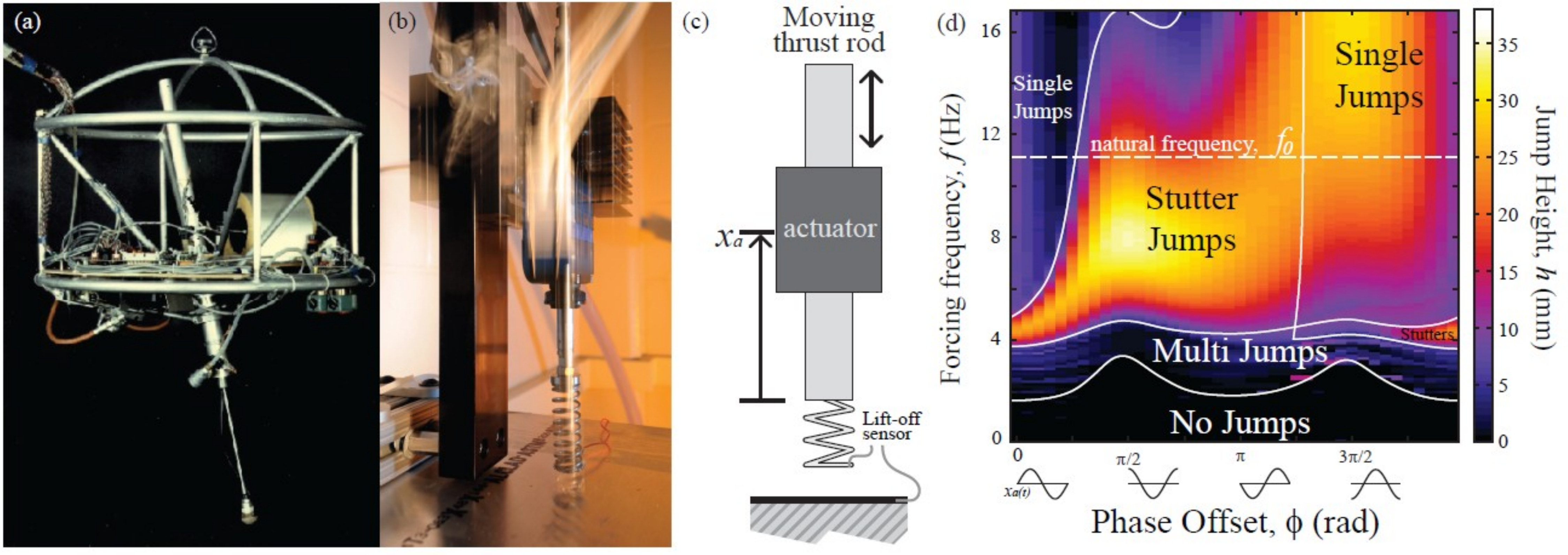}
\caption{\label{fig:jump} Hard ground jumping. (a) SLIP-inspired jumping robot~\cite{raibert1984experiments}. (b,c) Picture and diagram of simple jumping robot. (d) Color map of experimental jump heights vs forcing frequency and phase offset during 1 cycle sine-wave forcing. Approximately 20,000 experiments are represented. White lines separate regions of different jump types. (c-d) Adapted from~\cite{aguilar2012lift}.}
\end{centering}
\end{figure}

The intersection of robotics and physics has illuminated the surprising richness and complexity that robotic systems can exhibit, even in seemingly ``simple'' environments like rigid, flat, frictional ground. As an example, consider a robot hopping on hard ground. Many such devices have been created based on the spring-loaded inverted pendulum (SLIP) model, which broadly characterizes the dynamics of biological locomotor gaits such as running and hopping \cite{LABid2648697,blickhan1989spring,blickhan1993similarity,Blickhan2007}. Notably, Marc Raibert at MIT implemented the first SLIP-based hopping robot \cite{Raibert.Book1986}, paving the way to the creation and research of many more hoppers \cite{Akinfiev2003,Ahmadi1999,Armour2007,Brown,cham2007dynamic,German2000,Hayashi2001,matsuoka1980mechanical,Prosser1992,Peck2001,Mehrandezh1995,Okubo1996,Ringrose1997,Sato2004,Sayyad2007,Scarfogliero2007,Stoeter2002,Takeuchi2002,Tsukagoshi2005,Uno2002,Wei2000}. Most of these were made with an engineering focus on optimization \cite{German2000,Mehrandezh1995,Takeuchi2002}. Raibert in particular tackled the challenge of balancing a hopper able to move in 3D~\cite{raibert1984experiments} (Fig.~\ref{fig:jump}a). Raibert's robots have led to creation of arguably the most robust and successful real-world robots to date.

In our own research, we sought to gain a fundamental understanding of the dynamics of various jumping strategies, Aguilar et al.~\cite{aguilar2012lift} opted for the robophysics approach, and performed systematic experiments on a simplified 1-D SLIP hopper (Fig.~\ref{fig:jump}b,c). Balance issues were avoided by constraining the robot with a vertical air bearing. The actuation strategy of the motor was systematically varied by using a simple sinusoidal control template and adjusting parameters such as phasing and forcing frequency. The robot's simplicity facilitated the complete automation of the experiment, allowing for the performing and kinematic data collection of 20,160 jumps over a periods of days (Fig.~\ref{fig:jump}d). Two stereotyped jumping strategies emerged: the single jump, characterized by a single push-off, and the stutter-jump consisting of a preliminary hop followed by a push-off. A simple analytical model of the system (Fig.~\ref{fig:jump}b) revealed how optimal timing parameters of such maneuvers were influenced by resonance, transient dynamics and hybrid dynamic transitions between aerial and ground phases (particularly in the case of stutter jumps). Examining the simplest form the SLIP model revealed rich dynamics that are intrinsic to jumping performance, lessons that would have proved challenging to learn with more complicated robots.

Other robotic studies have taken similar approaches in analyzing SLIP-based robots. A common aspect among these studies is the systematic experimentation on simple robots to examine how various physical and actuation parameters produce and affect qualitatively different forms of hopping and jumping. In similar fashion to Aguilar et al.'s study, Senoo et al. \cite{senoo2010jumping} compared the performance of dynamically different jumping strategies, namely ``one-step'' and ``two-step'' jumping and examined the effect of bending angle on jump height on a simple 1-DOF jumper. For SLIP-based hopping, one early study by Koditschek and B\"{u}hler uncovered unique actuation solutions for stable hopping, one of which produced a two-period ``limping'' gait when using a nonlinear spring \cite{Koditschek1991}. Another hopping study further followed a robophysics format by not only systematically experimenting on a simple robot, but also performing a theoretical analysis using the tools from the field of dynamical systems. Varying body mass and stride period, Cham et al. \cite{cham2007dynamic} examined the stability of periodic hopping orbits during open loop actuation using the Poincare Map and Jacobian of the SLIP model and experimental kinematic data of the hopping robot. 
 
Another mode of interest to a number of roboticists is bipedal walking. However, even on hard ground, such a task can be difficult. As evidenced in the 2015 DARPA Robotics challenge~\cite{fallon2015architecture}, hard ground scenarios can contain complex obstacle arrays which challenge bipedal robots when performing tasks deemed simple for humans, such as opening a door. Typically, a motion planner analyzes sensory information to identify optimized obstacle-free motion trajectories within which to operate \cite{deits2015computing}. Without human assisted control, identifying these regions can be a challenge and sensitive to obstacle type, sensor uncertainty and deviations from the assumed model, often leading to failures.

Bipedal walking robots tend to perform better in simpler hard ground scenarios than in the above-mentioned unconstrained obstacle-laden environments, such as with the ATRIAS (Assume The Robot Is A Sphere) 2.1, an underactuated, planerized bipedal robot that uses the SLIP model to optimize walking gaits and minimize the cost of mechanical transport~\cite{ramezani2014}. Much like the air bearing constrained 1-D SLIP hopper~\cite{aguilar2012lift}, such constrained and simple environments facilitated a systematic robophysics-like examination of stable bipedal walking. The fundamentals of various pendulum-based walking models were to examined to create control algorithms for dynamically stable walking~\cite{kagami2002,chevallereau2009,grizzle2014, kuo1999, colArui2005} and induce passive dynamic walking down an inclined surface \cite{rushdi2013, spong2005, wisse2001}. Understanding these dynamic models to take advantage of passive elements has allowed for the creation of highly efficient robot walkers such as a bipedal robot that travels 65 km on a single battery charge \cite{bhounsule2014}. However, bipedal locomotion in more complex hard ground environments with obstacles will require an understanding of how the dynamics of grasping, manipulation, balance and sensing conspire to produce various successes and failures.

\begin{figure}[t]
\begin{centering}
\includegraphics[width={1\hsize}]{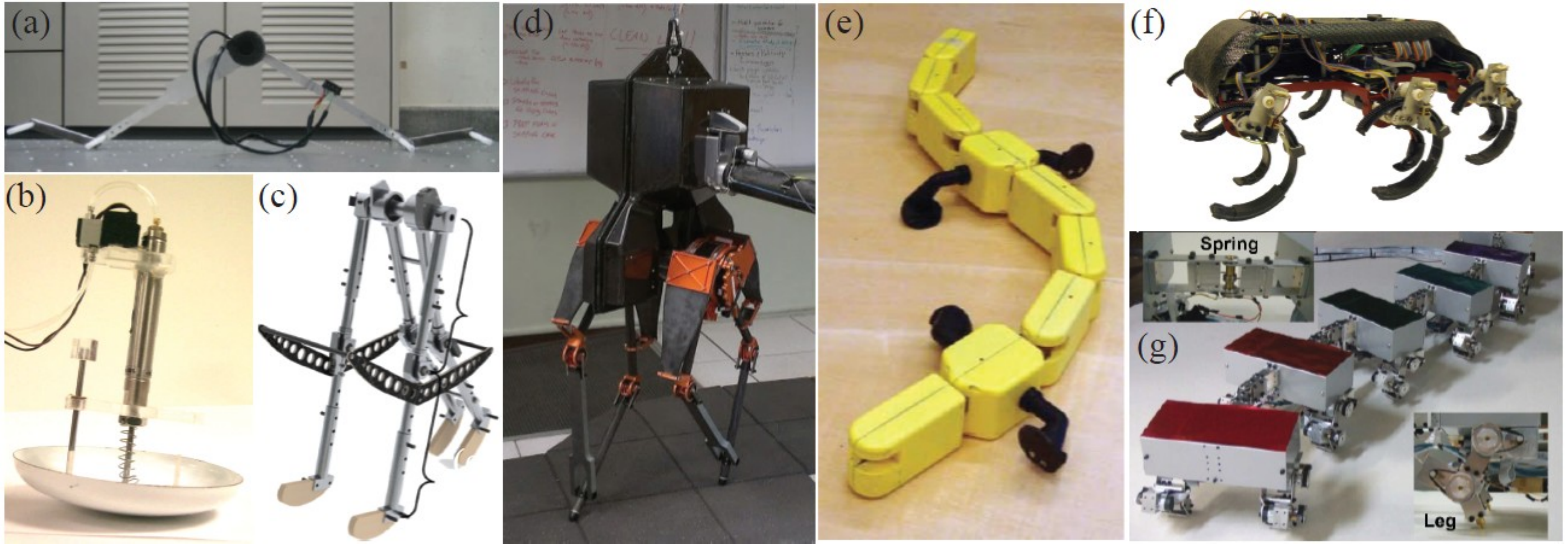}
\caption{\label{fig:hardground} Robot locomotion on hard ground. (a) 1-DOF jumper~\cite{senoo2010jumping}. (b) Dashpod~\cite{cham2001see}. (c) Passive walker~\cite{rushdi2013}. (d) ATRIAS~\cite{ramezani2014}. (e) Salamander robot~\cite{Ijspeert2007}. (f) RHex.~\cite{saranli2001rhex}. (g) Centipede-inspired robot~\cite{Aoi2013}.}
\end{centering}
\end{figure}

While other legged terrestrial devices with a greater number of legs lend themselves to higher model complexity, researchers have been able to establish systematic parameter exploration protocols to uncover locomotion principles. Such increases in complexity further highlight the necessity for robotic experimentation. Systematically changing robot and actuation parameters such as foot size~\cite{Qian2015BIBM}, joint angle, gait frequency~\cite{haldane2013animal}, and body undulation amplitude~\cite{Aoi2013} have been used to uncover principles relevant to locomotion such as contact physics between terrestrial locomotor foot and ground~\cite{Ding2013,kimAcla06}, and novel mechanisms for robotic locomotion such as anisotropic leg friction~\cite{leetraction,spagna07}, steering and forward movement with a single actuator~\cite{zarroukcontrolled}, scaled-impulse galloping~\cite{park2015quadrupedal}, and various mechanisms that enhance climbing~\cite{miller2015running,kellarrise,clarkclimbing2015,Kimclimbing2008}. Aoi's et al.'s studies on legged locomotion~\cite{Aoi2013,Aoi2007,Aoi2013a} are exemplary. These researchers used nonlinear dynamics and systematic variation of parameters to learn about movement on hard ground. Using a centipede-like robot as a physical model, they examined the amplitude and phase lag of body undulation at various locomotor speeds, and found that the undulation amplitude increased with increasing locomotion speed due to an instability caused by a supercritical Hopf bifurcation. A similar robophysics approach has also been used to study locomotor transition through different environments. In 2007, Ijspeert et al. studied the CPG pattern using a salamander robot driven by a spinal cord model, and analyzed the gait transition from aquatic to terrestrial locomotion~\cite{Ijspeert2007}. Since a number of these robotic studies take inspiration from biology, such studies have consequently been effective in examining their biological connections, from testing hypotheses of biological locomotor strategies~\cite{Aoi2013,kimAcla06,koditschek2004mechanical} to conducting a comparative study between robots and animals~\cite{Schroer2004}. \\

\section{Robophysics in air \& water}
\label{sec:fluid}

The power of modern day computers has revolutionized the ability to analyze the fluid interactions of locomotors through the simulation of Navier Stokes equations. However, understanding has likewise been advanced by the study of experimental robots. Here we give an admittedly abbreviated overview of examples of robot models that have advanced the study of locomotion in fluids; we apologize in advance to researchers whose work we do not cite. That said, robots have been used to study fish-like propulsion mechanics in water since the 1990's~\cite{triantafyllou1995efficient} and have been continuously developed and explored as mechanical models for swimming in water~\cite{Anderson01022002,Stefanini2012,Liu201035,Kato2000,Leftwich01022012,Sefati19112013,Esposito01012012} and flying in air~\cite{Shang2009}. Anderson et al.~\cite{Anderson01022002} investigated the swimming speed and turning ratio of the Draper Laboratory Vorticity Control Unmanned Undersea Vehicle (VCUUV), which utilized an early experimental design of the MIT RoboTuna~\cite{triantafyllou1995efficient}. Esposito et al.~\cite{Esposito01012012} designed a robotic fish caudal fin and systematically tested the effect of fin compliance and motion frequency on thrust generation. Groups led by MacIver~\cite{Sefati19112013,Cure11b,Neve14a,Bale15a,Neve13a}, Cowan~\cite{Sefati19112013} and Lauder~\cite{Curet2010} systematically studied swimming maneuverability and stability using undulating ribbon fin. Such advances have been built upon a rich history of study of relevant fluid-structure interactions~\cite{childress1981mechanics,taylor1952analysis,sfakiotakis1999review,sane2002aerodynamic}.

In recent years, there has been major increase in the use of unmanned aerial vehicles (UAV's), or drones. Applications include military reconnaissance and combat, crop monitoring, and very soon, package delivery. The decreasing cost of sensing and actuation technologies required for control of drones has facilitated the introduction of affordable consumer grade (and scale) aerial robots, suitable for low cost aerial footage~\cite{G.Loianno2015a}. Most relevant to our article, taking inspiration from biology, insect flight provides cautionary lessons by way of the deep issues in fluid dynamics needed to understand how such organisms (and small scale aerial robots) stay aloft~\cite{weis1973quick,dickinson1999wing,ristroph2010discovering}. One of the first detailed experimental studies of relevant forces was by Dickinson's group who used ``Robofly'' to discover principles of insect flapping flight and hovering~\cite{dickinson1999wing}. Wood's group designed the world's smallest flying robot~\cite{Shang2009} and explored the potential effects of wing morphology on flight performance. Rose and Fearing developed an empirical force model for flapping-winged fliers through systematic wind tunnel measurements of a flying robot, measuring elevator deflection and force for different wind speeds and angles of attack~\cite{rose2014comparison}. A benefit of these robot systems is the ability to model their interactions with fluids through the use of computation fluid dynamics and Navier-Stokes equations. Regardless, experimentation with these robotic systems has advanced the understanding of movement through these environments, because mere knowledge of Navier Stokes equations is insufficient to understand the importance of the interactions that emerge during various modes of locomotion between the fluid and dynamic locomotor models such as leading edge vortex structures during insect flight. Advances will continue to be aided by improved understanding of relevant aerodynamic interactions~\cite{wang2005dissecting}, as well as novel control schemes and designs~\cite{conroy2009implementation,saska:2014,ma2013controlled,cory2008experiments}. 

\begin{figure}[h]
\includegraphics[width=\textwidth]{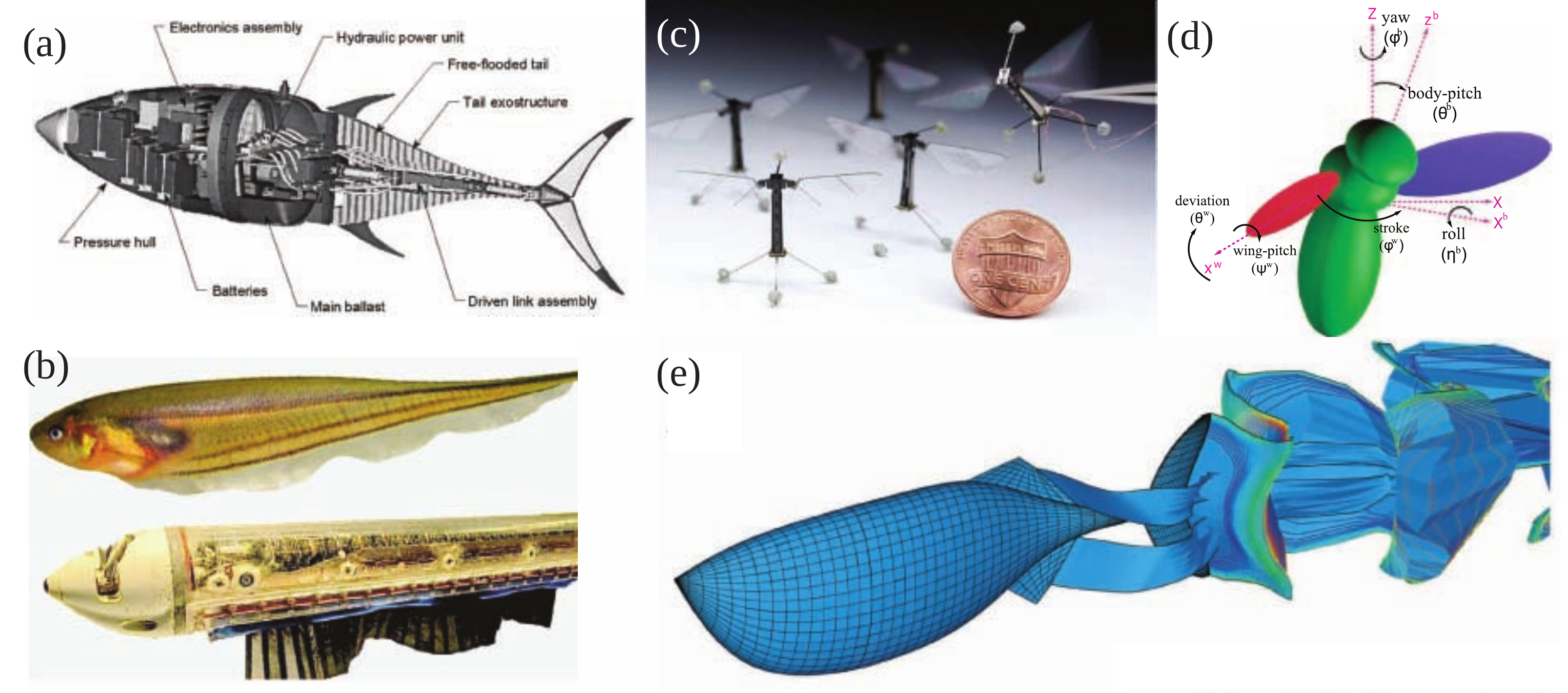}
\caption{\label{fig:figrobsim} Examples of robots that move through fluids. (a) Robotic Tuna~\cite{Anderson01022002}. (b) A Biomimetic robot for glass knifefish~\cite{Sefati19112013}. (c) Miniature flapping robot~\cite{ma2013controlled}. (d) Fluid simulation of insect flight \cite{Chang05082014}. (e) CFD (computational fluid dynamics) simulation model of a giant danio~\cite{Wolfgang01091999}.}
\end{figure}

\section{Robophysics in granular media, movement on and within a substrate between solid and fluid}
\label{sec:gm_robo}

We now discuss the environments which have occupied much of the locomotion robophysics efforts of our group: granular media.  We will demonstrate that dry granular media is an excellent system to study locomotion on ``flowable'' soft matter substrates. Dry grains displays a rich variety of behaviors in response to forcing and occur in many environments (deserts, dry regions of beaches) and regimes (small inertial number, see discussion in~\cite{andreotti2013granular}) encountered by robots. Such properties (especially homogeneity and lack of moisture) allow for precise experimental controllability of granular states. Further, dry granular materials are relatively simple to model accurately in numerical simulation: they can be described as an ensemble of individual particles whose emergent interactions are dominated by repulsive dissipative contact forces. Thus, modeling such interactions require only accounting for colliding spheres (and ``body'' elements which can be represented as spheres or flat surfaces), typically simpler to code and faster to solve than partial differential equations. We will first discuss a few studies which exemplify how the robophysics approach has proven critical to progressing our understanding of locomotion in granular media. Later in the review (Section~\ref{sec:compu}), we will discuss our modeling efforts.

\subsection{Legged locomotion}
\label{sec:rhex}
\begin{figure}[h]
\begin{centering}
\includegraphics[width={1\hsize}]{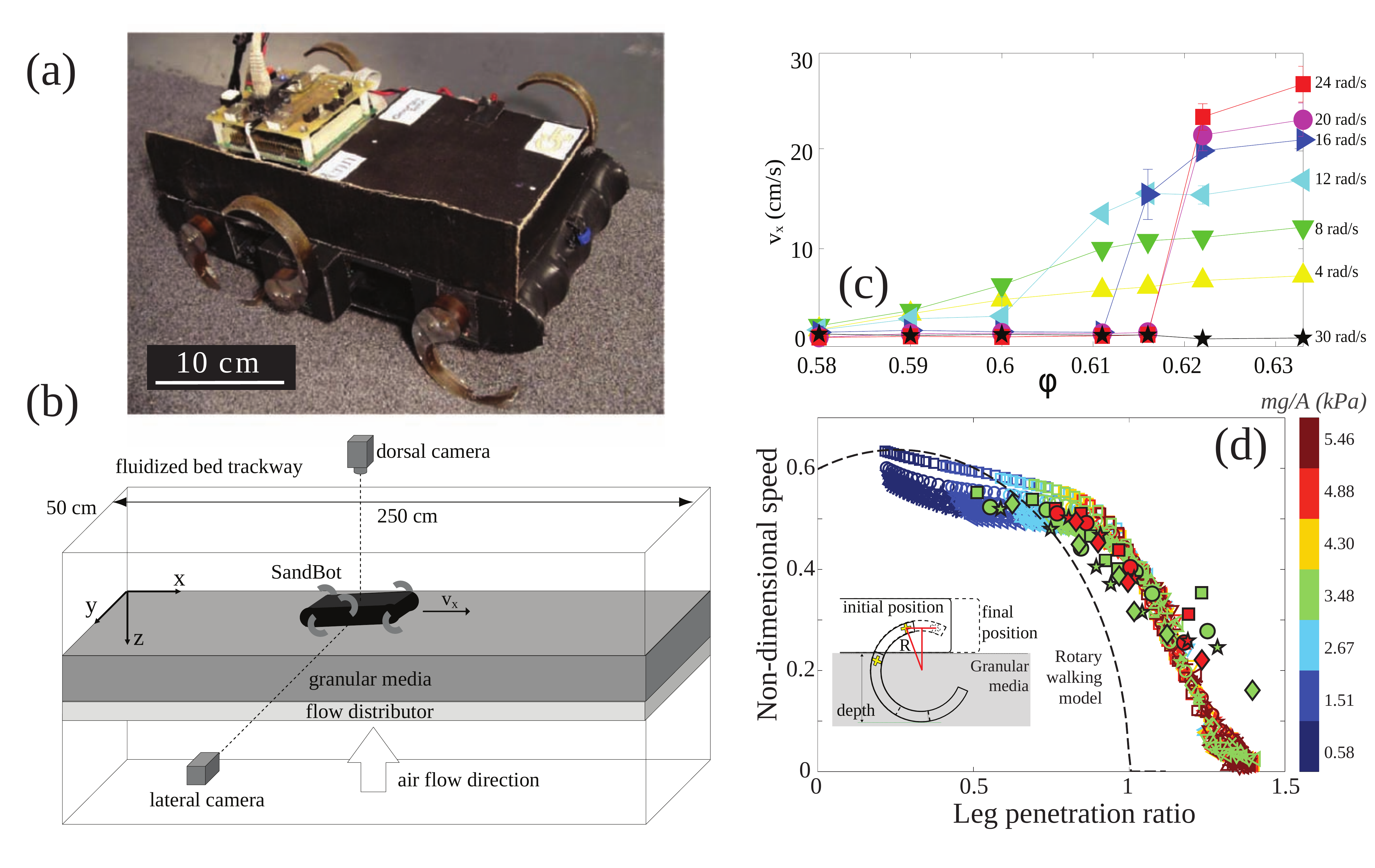}
\caption{\label{fig:Sandbot} Locomotion of a legged robot on granular media is sensitive to ground compaction and leg frequency. (a) The six-legged robot, SandBot, moves with an alternating tripod gait. (b) Ground control. Large flow of air followed by air pulses through the bottom of the fluidized bed trackway sets the initial volume fraction, $\phi$, of the granular substrate; air is turned off before the robot begins to move. (c) Forward robot speed is remarkably sensitive to $\phi$ at various limb frequencies, $\omega$. (d) Universal scaling of SandBot performance. Dashed line in diagram indicates rotary walking model. Dimensionless average forward speed vs.\ dimensionless leg insertion depth. Filled markers are experimental data for 4 gait frequencies, 2 foot sizes and 7 ground stiffness, and unfilled markers are terradynamic simulation for 10 gait frequencies, 7 masses and 20 ground stiffness. Marker shape indicates gait frequency ($\Box$: $2~{\rm rad~s^{-1}}$; $\circ$: $4~{\rm rad~s^{-1}}$; $\bigstar$: $6~{\rm rad~s^{-1}}$; $\lozenge$: $8~{\rm rad~s^{-1}}$; $\Cross$: $10~{\rm rad~s^{-1}}$; $+$: $12~{\rm rad~s^{-1}}$; $\vartriangle$: $14~{\rm rad~s^{-1}}$; $*$: $16~{\rm rad~s^{-1}}$; $\vartriangleleft$: $18~{\rm rad~s^{-1}}$; $\vartriangleright$: $20~{\rm rad~s^{-1}}$; Color indicates the ratio of body weight to foot size as shown in colorbar. Adapted from~\cite{li2009sensitive} and ~\cite{Qian2015BIBM}.}
\end{centering}
\end{figure}

Robophysical devices have proven especially useful for examining contact dynamics during legged locomotion on dry granular media. Li et al. systematically studied the locomotion of a RHex-class robot (Sandbot), a legged device that is tuned to run on hard ground~\cite{saranli2001rhex}, and discovered that ground reaction forces and robot dynamics on granular media depended sensitively on actuation parameters such as leg frequency~\cite{li2009sensitive} and intra-cycle leg kinematics (relative phasing between fast and slow leg rotations)~\cite{Li2010}. Using SandBot as a physical model, Li et al. found that the robot kinematics at low leg frequency could be described using a rotary walking model~\cite{li2009sensitive} (Fig.~\ref{fig:Sandbot}d inset) which could predict robot step length and forward speed. According to this model, the legs penetrated a yielding substrate until granular reaction forces balanced robot weight and body inertia, at which point the substrate solidified and the legs stopped penetrating and began rotating to propel the body forward.\\

In recent work, Qian et al.~\cite{Qian2015BIBM} demonstrated that the rotary walking model could be generalized to explain the locomotion performance for both robotic and biological locomotors with scale from 10 g to 2.5 kg. By systematically studying locomotor performance (i.e., average forward speed) of a 2.5 kg, cylindrical legged robot, Qian et al. derived a universal scaling model (Fig.~\ref{fig:Sandbot}d) that revealed a sensitive dependence of speed on the leg penetration ratio for all stiffnesses and gait frequencies. The model was then applied to running lizards, geckos and crabs, to explain the differences in performance loss observed in these animals as ground stiffness was reduced. To extend our result to include continuous variation of locomotor foot pressure, we also applied RFT to perform numerical simulations, and found that the RFT prediction agreed well with experiments for various robot sizes as well as various granular substrate stiffness. Despite the variation in morphology and gait, the performance of lizards, geckos and crabs were also determined by their leg penetration ratio, as the universal model predicts. A further analysis of performance loss rate suggested that locomotors with smaller foot pressure can passively maintain minimal leg penetration ratio as the ground weakens, and consequently permit effective high-speed running over low stiffness ground. To summarize, this legged robot moves effectively on granular media when design and actuation strategy minimize yielding of the substrate.

\subsection{Sidewinding}
\label{sec:sidew}
\begin{figure}[h]
\begin{centering}
\includegraphics[width={1\hsize}]{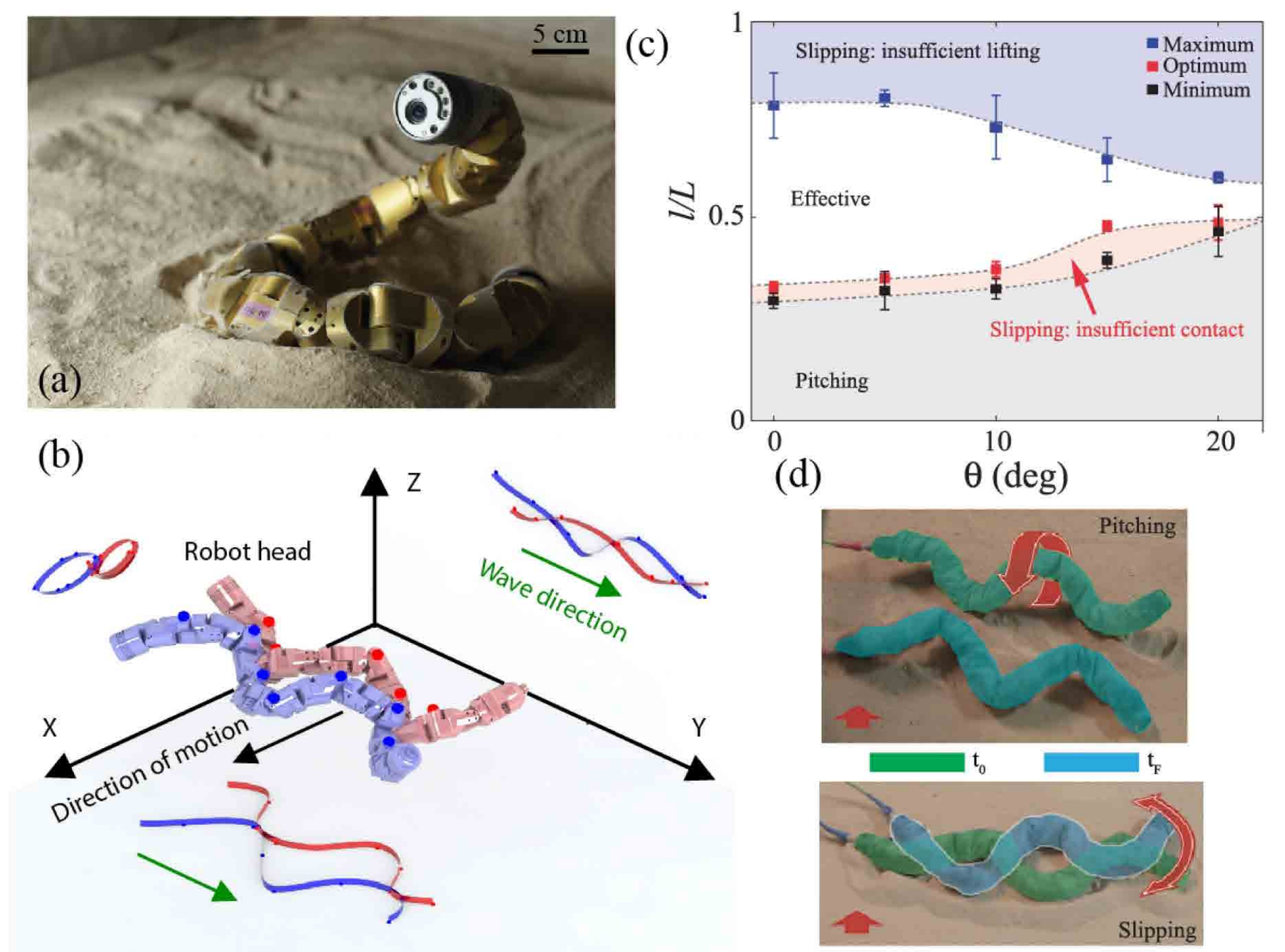}
\caption{\label{fig:mod_snake} Robotic sidewinding. (a) The CMU modular snake robot on sand. (b) Snapshots of the CMU snake robot executing a sidewinding gait on level sand and projections of the resultant body shapes onto three orthogonal planes. Red and blue indicate the initial and successive images, respectively. Time between images is 6.3 s. The horizontal and vertical waves travel in the posterior direction of the robot with respect to a body-fixed coordinate system (as shown by the green arrows). (c) Speed versus l/L vs inclination angle, $\theta$, with colored regions indicating failure regimes due to pitching and slipping. Three trials were performed at each condition. Data indicate mean $\pm$ SD. (d) Super-imposed frames showing pitching and slipping failure modes in the robot ascending $\theta = 20^{\circ}$ and $10^{\circ}$ inclines, respectively. Uphill direction is vertically aligned with the page. $t_{0}$ and $t_{F}$ represent the time at which each body configuration is captured. The time between two images in the pitching and slipping failure modes is 1.6 and 6.3 s, respectively. Adapted from~\cite{marvi2014sidewinding}.}
\end{centering}
\end{figure}

Limbless locomotion on dry granular media is another excellent system for robophysical study. In particular, the sidewinding gait exemplifies how robophysics can aid physics, engineering and biology. Sidewinding is a strange gait observed only in certain snakes that live on flowable material, such as the rattlesnake (\emph{Crotalus cerastes}) and prescribed by engineers to move limbless robots over diverse terrain. Kinematic data from biological sidewinders is relatively easy to acquire and provides a strong basis on which to ground this study. However, the lack of available computational models capturing the snakes dynamics on sand makes it difficult to systematically test hypotheses formulated from biological data. This highlights the need for a controllable physical model that approximates the biological system's motion and environmental interaction. An example of such a model is the modular snake robot locomoting on granular media used in~\cite{marvi2014sidewinding}.

The snake robot in~\cite{marvi2014sidewinding} (Fig.~\ref{fig:mod_snake}a) was used to study the physics of sidewinding on granular inclines. Experimental kinematic data of (\emph{Crotalus cerastes}) sidewinding revealed an increase in length of body segments in contact with the sand as the inclination of the granular slope increased. It was hypothesized that snakes regulate contact length to minimize slip while traversing sloped surfaces. To test this, sidewinding motion was implemented on the snake robot using a two-wave model~\cite{marvi2014sidewinding,Henry_PNAS} comprised of orthogonal waves posteriorly traveling in the horizontal and vertical body planes (Fig.~\ref{fig:mod_snake}b). Modulation of the relative amplitudes of the two waves changed the eccentricity of the sidewinding robot's overall shape, which in turn regulated the length of contacting body segments. Testing contact lengths on different slopes revealed two distinct failure modes (Fig.~\ref{fig:mod_snake}d): tipping over due to gravity, or slipping due to the substrate yielding. As the inclination of the surface increased, the robot had to increase contact length to both maintain balance as well as reduce the stress applied to the substrate. However, excessive contact length (insufficient lifting) introduced unnecessary drag force which in turn caused the substrate to yield. Thus, using grid-based optimization, it was empirically determined that an optimal length of contact as a function of the slope existed, one which maximized stability while minimizing slip and drag. This observation confirmed the importance of biological sidewinders regulating contact length to minimize slip/yielding on granular slopes.

This result led to a larger idea: the sidewinding gait is a control template (a neuromechanical target of control) for locomotion on granular material, one that allows effective locomotion on yielding granular media through appropriate manipulation of the substrate. The template is formed when the snake (or the robot) propagates two waves down the body, one in the horizontal plane, the other in the vertical plane (as discussed above). In both the animal and robot, these waves are phased by $\pi/2$ radians. The increase in contact length can be viewed as an amplitude modulation of the vertical wave to control for possible yielding of the granular material. Other modulations of the two-wave template produce turning behaviors in the robot similar to those observed in the animals; in ~\cite{Henry_PNAS}, we discovered that the snakes could reorient over several gait cycles (which we called ``differential turns'') or could rapidly (within a cycle) change direction (which we called ``reversals''). Neither of these turns produced significant slip or flowing of the material. In the robot, modulation of the horizontal wave with increasing amplitude from head to tail resulted in differential turns while a sudden phase modulation of the vertical wave by $\pi$ radians yielded reversal turns with performance comparable to the animal. More complex modulations of the waves, such varying the relative spatial frequency between the two waves (``frequency turning''), yielded turning behaviors not observed in the animals. These modulations exemplify how robophysics can be used to anchor template models of locomotion.

\subsection{Flipper-based terrestrial locomotion}
\label{sec:flip}
\begin{figure}[t]
\includegraphics[width=\textwidth]{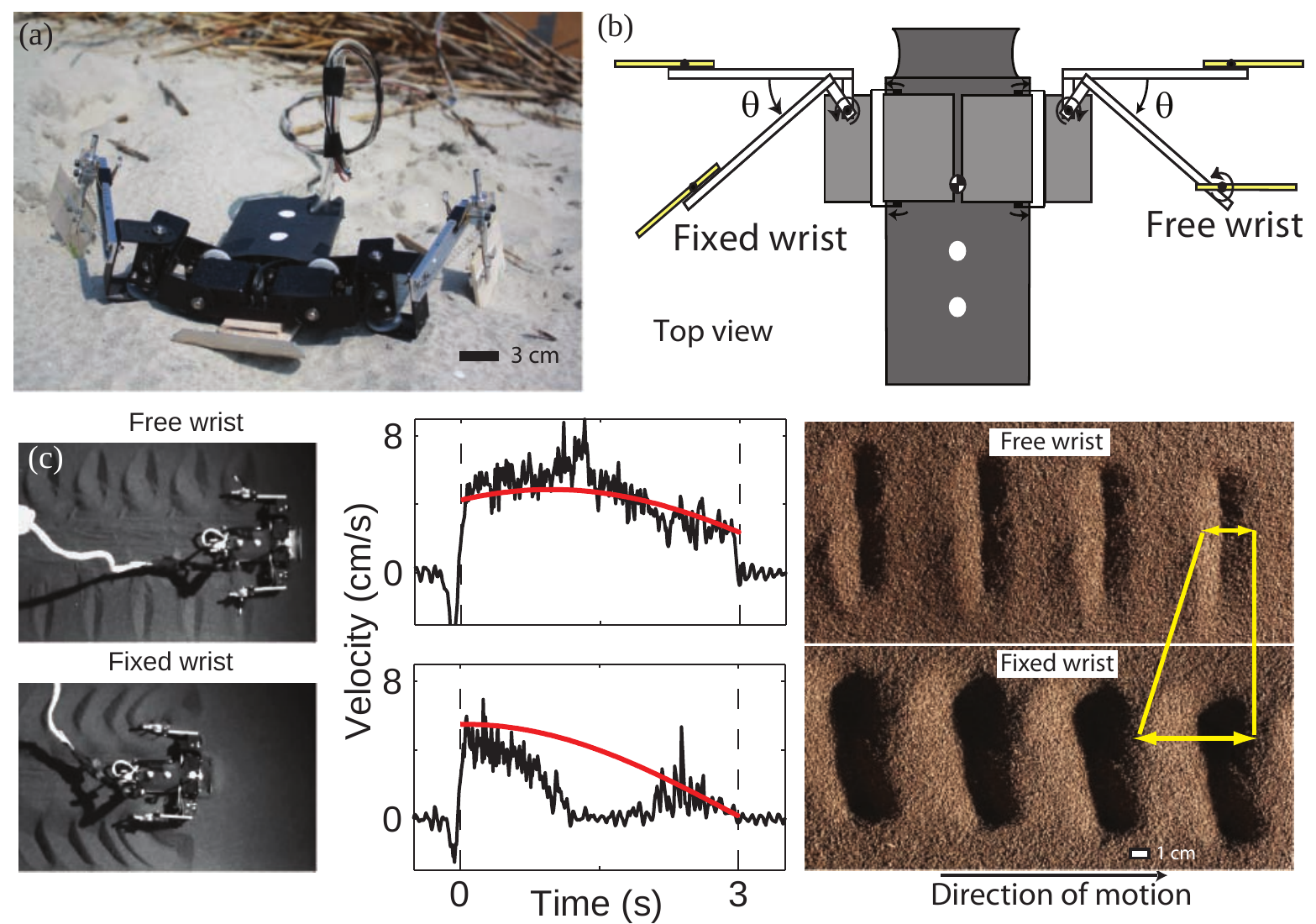}
\caption{\label{fig:FlipFig3} Flipper-based terrestrial locomotion. (a) Sea-turtle inspired FlipperBot. (b) Diagram indicating fixed (left) and flexible (right) wrist flipper configurations. The robot inserts flippers into media to some depth and then strokes from front to back through an angle, $\theta=\pi/2$. (c) Comparison of forward displacement of FlipperBot with a flexible wrist (top) and rigid wrist (bottom). Experimental (black) vs theoretical (red) velocity vs time in center panel. Fixed wrist model induces a granular frictional fluid which produces a larger area of granular disturbance than free wrist (right panel), thus increasing the amount of weakened substrate that subsequent strokes interact with. The free wrist model is kinematic and flipper stays below granular yield. Adapted from~\cite{mazouchova2013flipper}.}
\end{figure}

Flipper-based terrestrial locomotion (like that used by sea turtles) highlights the challenge of using aquatically adapted appendages to locomote on land. Understanding this mode of locomotion has benefited from a robophysics approach, which illuminates the influence of mechanical design on the induction of qualitatively different substrate behaviors, in this case generating solid states of granular media to improve locomotor performance  and reduce suspectibilty to failure. Our biological field observations of Loggerhead hatchlings~\cite{mazouchova2010utilization} revealed that the hatchlings utilize granular solidification by bending their wrist during limb-ground interaction with sand, and employ a rigid wrist while walking on hard ground, allowing for comparable forward speeds on both surfaces. To understand how such limb-ground interactions affect locomotion performance we created FlipperBot (Fig.~\ref{fig:FlipFig3}a), a servo-motor-driven robot which propels itself using limbs and flat-plate flippers \cite{mazouchova2013flipper}(Fig.~\ref{fig:FlipFig3}b).

Using a fixed wrist joint revealed that FlipperBot was prone to failure by interacting with ground disturbed by previous steps, resulting in a decrease of flipper generated forces as the per-step distance was reduced (Fig. \ref{fig:FlipFig3}c). Conversely, a flexible wrist made the robot less prone to failure since there was less interaction with previously disturbed ground (Fig. \ref{fig:FlipFig3}c). The use of FlipperBot additionally allowed for systematic changes in flipper insertion depth, showing how extremely shallow depths (1.4 mm) prevented forward movement. Overall, FlipperBot proved essential in understanding how kinematic control and structural design affect the dynamics of flipper-based locomotion on sand. This flexible limb use enables control of the state of the granular media. Given the difference in morphology between snakes and Flipperbot, this is remarkably similar to how sidewinders modulate body contact length to control the granular state.

\begin{figure}[th]
\begin{shaded}
\textbf{\begin{large}
Sidebar: robophysics applied to evolution of movement
\end{large}}\\

\begin{centering}
\includegraphics[width={.8\hsize}]{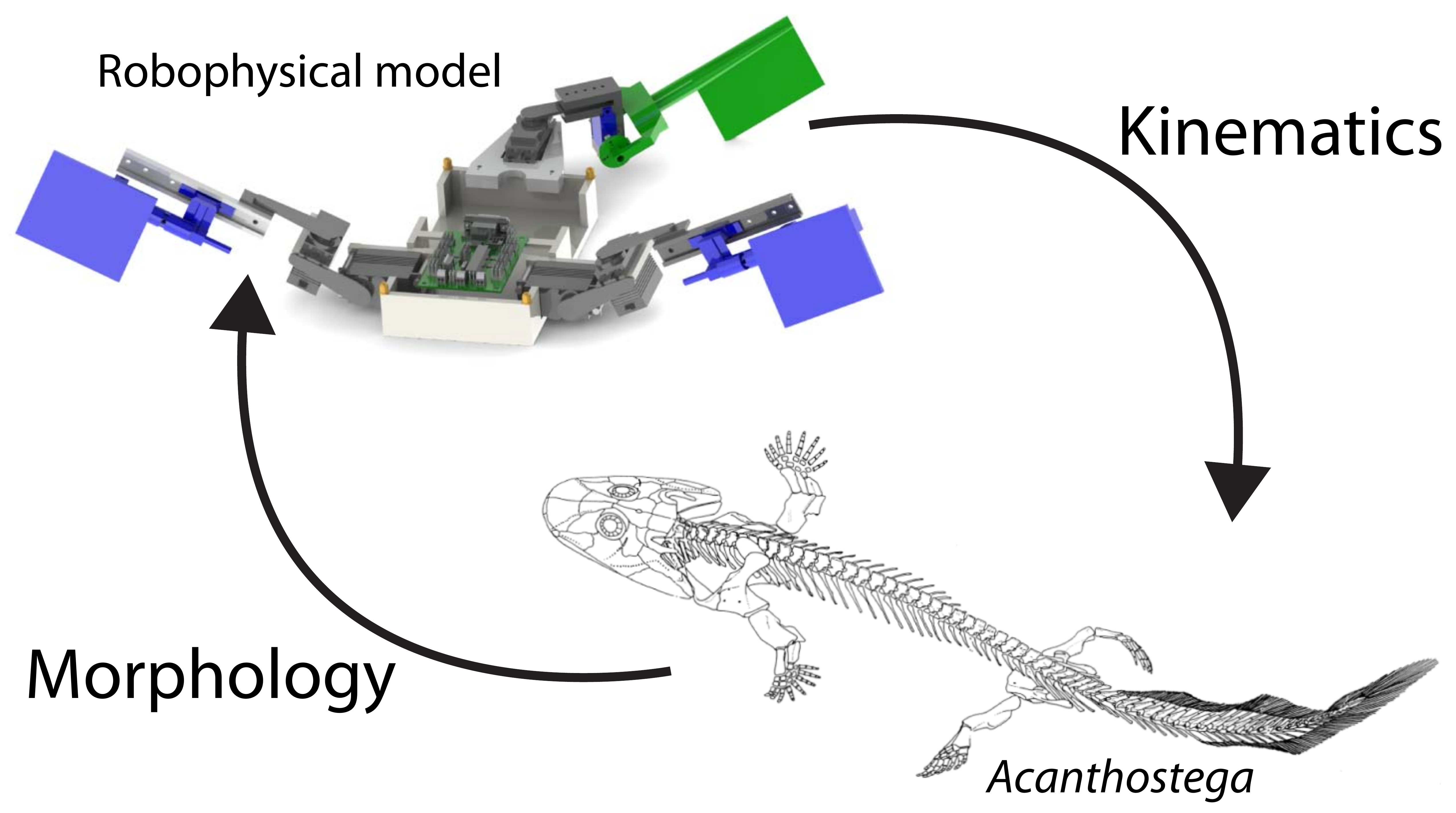}
\caption{\label{fig:muddy} \scriptsize{ A robot model of locomotion, the MuddyBot (left), morphologically inspired by skeletal reconstructions of early terrestrial walkers such as the \textit{Acanthostega}~\cite{Coates_acanth} (right), will bring insight into the physical mechanisms of locomotion of these extinct organisms.}}
\end{centering}

\bigskip

\scriptsize{
Robophysics not only allows for systematic study of the locomotion of living organisms, but also provides a framework for understanding the evolution of locomotion in extinct taxa, such as avian flight evolution~\cite{peterson2011wing}. Additionally, there exist skeletal reconstructions of early terrestrial walkers, such as Icthyostega\cite{Pierce_ichthy}, Tiktaalik\cite{Shubin_tik}, and Acanthostega\cite{Coates_acanth}, which provide insight into the limb-joint morphology of these organisms. However, previous robotic experiments\cite{li2009sensitive}\cite{mazouchova2013flipper} have shown that morphology alone is insufficient to determine locomotor efficacy. Previous biomechanical studies of early walker locomotion have relied on extant modern analogue organisms to gain insight into possible limb kinematic strategies.\cite{King_lungfish}\cite{Kawano_muddy}. However, without the ability to systematically vary kinematic parameters in the proper morphological context, the physical mechanisms of early walker locomotion have remained elusive. By developing a robot with limb-joint morphology inspired by early walker skeletons, and implementing kinematic strategies and behaviors observed in modern analogue organisms~\cite{McInroeMuddy}, we can bridge the gap between past and present fauna, extending the reach of the physics of living systems into the distant past.
}
\end{shaded}
\end{figure}

\subsection{Sand-Swimming}
\label{sec:sandswim}
\begin{figure}[htbp]
\includegraphics[width=\textwidth]{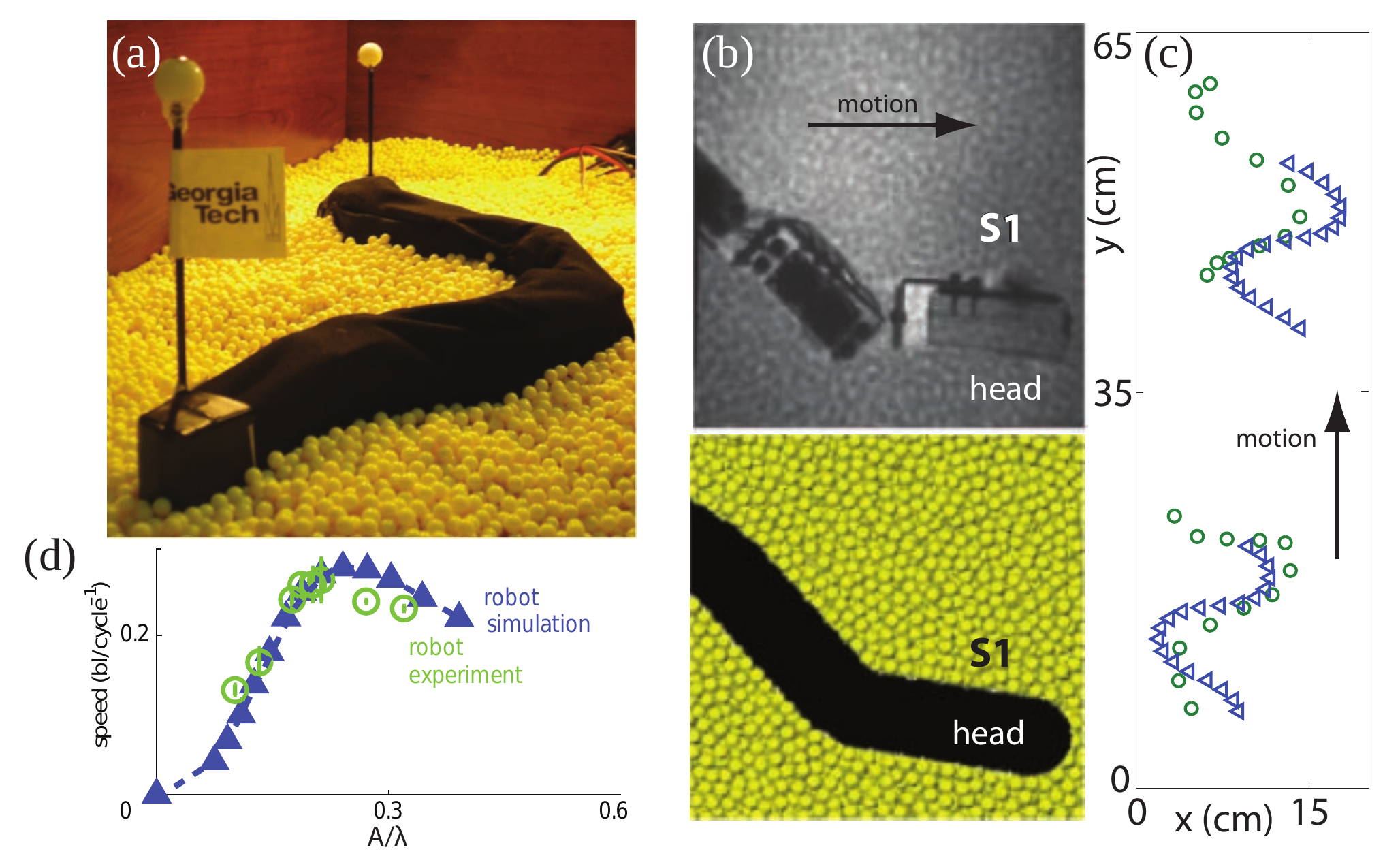}
\caption{\label{fig:figSS} Undulatory robot swimming in granular media. (a) a 7-link servo-motor actuated robot in a bed of 6 mm particles. (b) X-ray image of the sub-surface swimming of the 7-link robot in experiment (top) and cross section in DEM simulation (bottom). (c) trajectories of swimming in experiment (green circles) and simulation (blue triangles). (d) Body lengths per cycle as a function of wave amplitude for a single period wave. (a-d) adapted from~\cite{Maladen2011a}}
\end{figure}

Understanding the undulatory mechanics of sand swimming performance has also benefited from a robophysics approach~\cite{Maladen-RSS-10, maladen2011} in conjunction with simulation and biological observations~\cite{Maladen2009, ding2012mechanics,sandfishtemplate}. Inspired by subsurface swimming of the sandfish lizard (\textit{Scincus scincus})~\cite{Maladen2009}, Maladen et al. assembled a  7-link robot (Fig.~\ref{fig:figSS}a) that can perform lateral undulation inside granular media of 6 mm plastic particles (Fig.~\ref{fig:figSS}b, top panel)~\cite{Maladen-RSS-10, maladen2011}. The locomotion was also studied numerically: using a multi-body solver (WorkingModel), a ``virtual'' robot interacted with granular particles simulated by an experimentally validated DEM (discrete element method, see Section~\ref{sec:compu}) simulation (Fig.~\ref{fig:figSS}b, bottom panel). Agreement between simulation and experiment was observed (Fig.~\ref{fig:figSS}c,d, to be discussed further in Section~\ref{sec:highspeed}). Variation of undulation control parameters revealed that forward swimming speed was maximized with a single period body wave and wave amplitude, $A/\lambda\approx 0.2$ (Fig.~\ref{fig:figSS}d)~\cite{Maladen2011a}. Unlike the studies of locomotors on the surface of granular media which solidify the ground for effective push-off, in sand-swimming, the control target facilitates the creation of a localized granular ``frictional fluid'' which extends roughly a body-width away from swimmer~\cite{ding2012mechanics}.

\section{Heterogeneous terrestrial substrates}
\label{sec:hetsub}

\begin{figure}[h]
\begin{centering}
\includegraphics[width={1\hsize}]{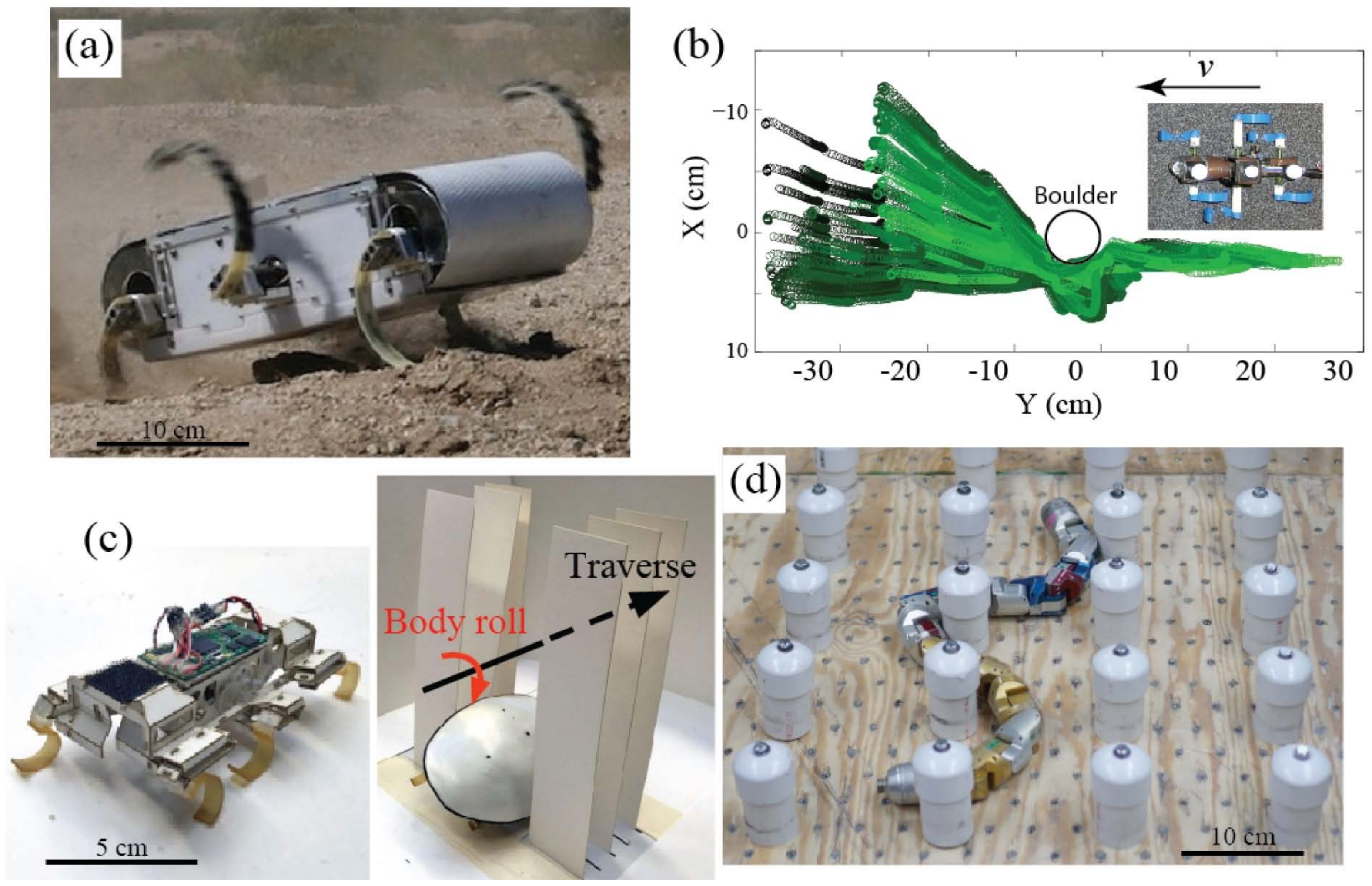}
\caption{\label{fig:hetfig} Robots in terrestrial heterogeneous environments. (a) RHex robot traveling across heterogeneous gravel substrate (photo courtesy Alfred Rizzi, Boston Dynamics). (b) Robot interaction with single spherical boulder. The center of the boulder is set as the origin. Robot (inset) CoM trajectories for the single boulder interaction. Trajectory colors represent variation in robot initial fore-aft positions, which influenced the angle of scatter caused by the boulder. (c) A small hexapedal robot negotiating densely cluttered, grass-like beams (adapted from ~\cite{li2015grass} with permission from Bob Full). Left: With its initial cuboidal body shape, the robot always changes heading after contact with the beam and becomes stuck. Right: By adding a thin, rounded shell inspired by the discoid cockroach, the robot traverses beam obstacles via body rolling, without adding sensors or changes in control. (d) A snake robot moving on a reconfigurable peg board. The configuration of the pegs can be modified by installing pegs at different locations on the board.}
\end{centering}
\end{figure}

The environments of the natural world contain materials of incredible complexity and often heterogeneity. Many animals are quite robust in their movement to these complexities. As an example, forest-dwelling insects encounter cluttered environments with various obstacles like grass, shrubs, trees slabs, mushrooms, and leaf litter. To understand the mechanisms for effective movement in these environments. Li et al. studied discoid-shaped cockroaches traversing a field of densely packed, grass-like beams and discovered that wearing artificial body shells of reduced roundness reduced their traversal performance~\cite{li2015grass}. They then examined how such changes in body roundness affected the traversal efficacy of a small, open-loop-controlled, legged robot, and observed a similar effect as found in the insects. A rounded cockroach-inspired shell enabled the robot to traverse densely cluttered beams via body rolling, as opposed to failure to traverse with the initial cuboidal body shape (Fig.~\ref{fig:hetfig}b). If robots of the future are to effectively maneuver across terrains of interest such as disaster sites, comets and extraterrestrial planets, we must understand the mechanisms of effective movement in such terrains. This will only be possible through the parametrization and systematic creation of heterogeneous granular ground. However, characterizing how the nearly limitless variability in size, shape, granular compaction, gap, orientation, etc. affects locomotion can seem like a daunting task to probe by hand. In fact, without automation, collecting the large statistical data sets that will likely comprise the type of analysis that robophysics provides will be nearly impossible. The following study exemplifies automation's utility in the study locomotion of heterogeneous substrates.

Qian et al. ~\cite{Qian2015RSS} have taken the first steps in this ambitious direction with the development of a fully-automated terrain creation system called Systematic Creation of Arbitrary Terrain and Testing of Exploratory Robots (SCATTER, Fig.~\ref{fig:auto}c). The SCATTER system consists of an air fluidized bed that sets sand compaction, two tilting actuators that control the substrate inclination, and a universal jamming gripper~\cite{Brown2010} that retrieves and re-distributes boulders and the robot after each locomotion test. Using the SCATTER system, properties of heterogeneous substrates such as compaction, orientation, obstacle shape/size/distribution, and obstacle mobility within the substrate, can be precisely controlled and varied. A high speed tracking system was also integrated into the SCATTER system, to track the robot kinematics in 3D, observe robot and boulder interaction, and locate robot and boulders after each test for gripper retrieval. SCATTER, which is high throughput like the above mentioned systems ($\sim200$ tests/day without human intervention) has already revealed important properties of open loop locomotor scattering of a small hexapedal robot (15 cm, 150 g) during various boulder interactions~\cite{SPIE2015}.  The SCATTER trackway was filled with fine granular media (the ``sand'') and a single ``boulder'' (3D printed convex objects of different geometries) embedded in. Analysis of the robot’s trajectory indicated that each interaction could be modeled as a scatterer with attractive and repulsive features, whose magnitude depended sensitively on the local boulder surface inclination at initial contact point. Depending on the contact position on the boulder, the robot will be scattered in different directions (Fig.~\ref{fig:hetfig}c). For a larger heterogeneous field with multiple boulders, the trajectory of the robot can be statistically estimated using a superposition of the scattering angle from each boulder. This scattering superposition can be applied to a variety of heterogeneity, including different geometry, orientation, and roughness. An analogy to the scattering problem simplified the characterization of the heterogeneous ground effect on robot trajectory deviation, and allowed for long term dynamics analysis for exploratory robot trajectories on large, complex heterogeneous fields. The design lessons learned from SCATTER will serve as a blueprint for the construction of more arbitrary terrain creation systems.

Choset\textsc{\char13}s group is exploring different control strategies to enable snake robots to effectively maneuver heterogeneous environments.  A reconfigurable peg board was constructed to allow systematic variation of environment geometry (Fig.~\ref{fig:hetfig}d). Peg configuration can be easily adjusted by installing pegs at different locations on the base. Two controller types were examined: position-based control and force-based control. The performance of the position-based controller, which generated effective motion on homogeneous substrates, quickly degraded in heterogeneous environments. The performance of this controller was sensitive to small perturbations of the control parameters. Two types of failure modes, backward slip and jamming, were observed in the robot experiments. Conversely, force-based control showed advantages over position-based control in generating more robust motion. Simulation of the snake robot on peg board revealed the ``geometric'' nature of the system, where the shape/geometry of the extended body coupled to the environment space. The performance of a position controlled snake is purely dominated by the ratio between its wavelength to peg distance. However, when the robot is force controlled, it can stretch or shrink to modulate its wavelength that better adapts to different peg spacings. Thus, while many irregularities are rigid and cannot be manipulated like granular media, effectively leveraging an environment's obstacles is crucial for successful locomotion.

\section{Wet terrestrial substrates}
\label{sec:wet}
Thus far we have focused on robophysical studies of movement in air, water, hard ground as well as on and within dry granular media. However, wet granular substrates present interesting locomotor challenges. Going from slightly wet to fully saturated creates dramatic changes in the rheology of such substrates. The complexity of structures that can form in granular media are related to moisture content, from smooth piles using dry grains, to sandcastles made possible by slightly wet sand, to slurries formed by fully saturated sand; for a comprehensive review of the physics of such states see~\cite{mitarai2006wet}.

\begin{figure}[h]
\begin{centering}
\includegraphics[width={1\hsize}]{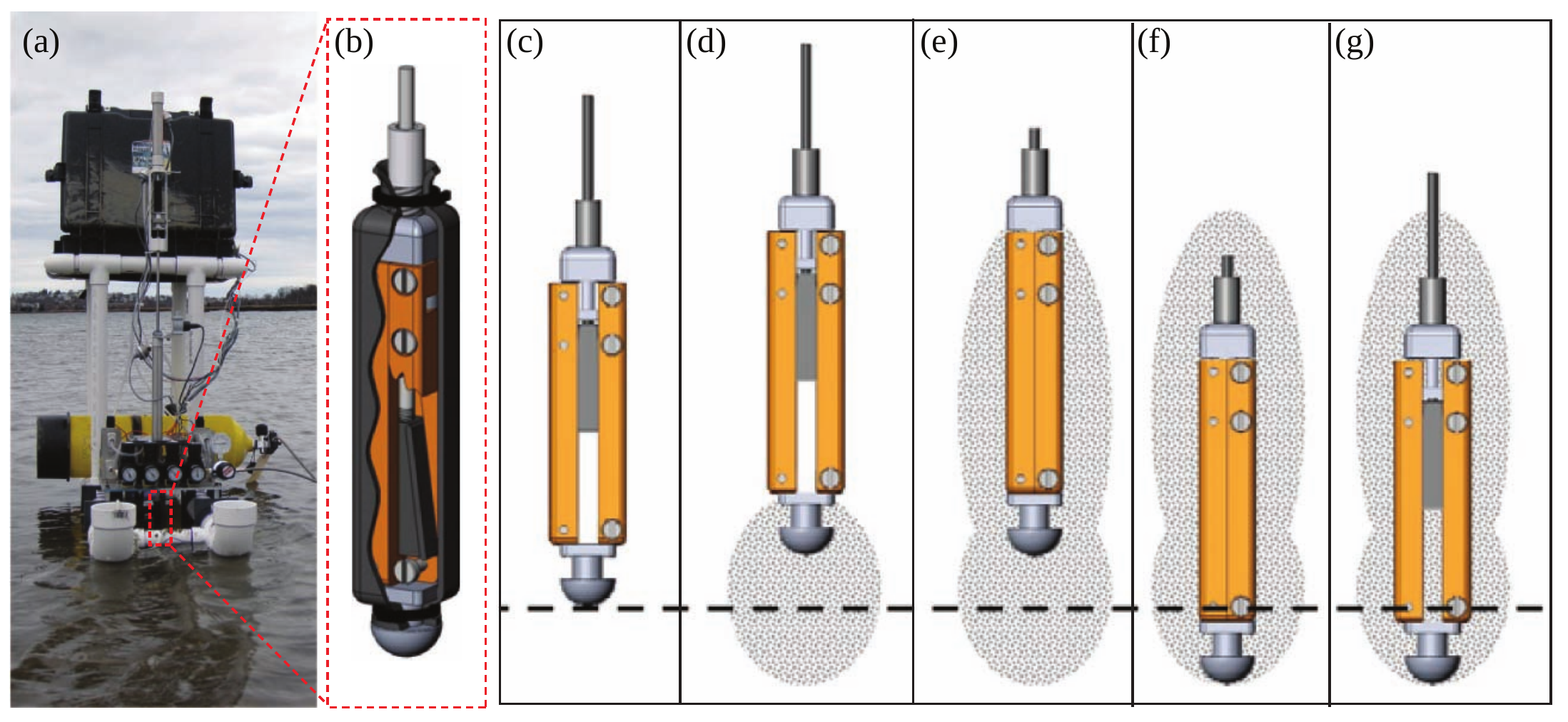}
\caption{\label{fig:roboclam} The RoboClam, a device which can dig in fully saturated soils using a two anchor burrowing technique with local fluidization. (a) The RoboClam on a mudflat, consisting of a scuba tank for compressed air actuation, pressure control valves, expansion-contraction and insertion-retraction pistons, a data acquisition and control laptop and an end-effector, (b), which is structurally inspired by the \textit{E. directus}. (c-g) illustrate the robot end-effector motions while burrowing, where the dashed line is a datum for depth, and the shaded grey areas indicate expected local fluidization of soil. The end-effector follows a sequence of retracting, contracting, inserting and expanding that fluidizes the substrate and weakens its resistive forces to facilitate burrowing. Figure adapted from Figure 5 in~\cite{winter2014razor}.}
\end{centering}
\end{figure}

An excellent robophysical study in fully saturated granular media was performed by a group led by Winter and Hosoi~\cite{winter2014razor,winter2012localized} in a regime of fully saturated granular media (Fig.~\ref{fig:roboclam}). This group performed studies on a robotic digger, the so-called ``RoboClam'', modeled on the Atlantic razor clam {\em Ensis directus}. This clam, which digs vertically in mudflats through a two-anchor method, is able to descend to depths in soil where penetration resistance far exceeds the capability of the organism to generate force. By creating a device that modeled the digging kinematics and mechanics of the organism, and then systematically varying parameters (as well as developing genetic algorithms to optimize digging), the group discovered how strategically inducing local fluidization during a phase in the burrowing cycle can decrease force requirements, thereby enabling the robot (and presumably the organism) to dig deeply into mud.

Many terrestrial soils are composed of wet substrates that are not fully saturated, and in such substrates it is challenging to create mixtures of granular media and water whose initial conditions can be set precisely and with uniform homogeneity. Recently, members of our group have developed a vibrational sieving technique to create laboratory containers of wet granular media of different moisture contents and compaction and have used this technique to study fire ant digging~\cite{Gravish2012fireant,gravish2013climbing} and lizard locomotion~\cite{sharpe2015controlled}. We anticipate a bright future for new questions in localized intrusion in such substrates. Interaction studies typically deal with boundary forcing of the entire ensemble of wet granular states (for a nice review see~\cite{mitarai2006wet}), whereas drag measurements in~\cite{sharpe2015controlled} have revealed complex dynamics which could inspire new soft matter physics for wet granular media. Robotic devices tasked with traversing through arbitrary substrates could benefit from a robophysics within such materials.

\section{Computational tools in locomotion robophysics}
\label{sec:compu}

While the above experiments serve as a kind of ``physical simulation'' of more complex robots and their biological counterparts, robophysics, like all modern sciences, greatly benefits from an integration of experiment and computational modeling tools. These tools have been useful in making quantitative predictions and explaining underlying mechanics. Computational fluid dynamics (CFD) has progressed the study of aerial and aquatic locomotion. Insights into hard ground locomotor interactions~\cite{saranli2001rhex} have also benefited from numerical simulations which incorporate methods of contact and collision, the inherently discrete nature of which has resulted in nontrivial challenges in accurate modeling.

Granular media is an example of a complex substrate that can behave like a solid, a fluid or a gas (the latter when an ensemble of granular media is sufficiently agitated, for a broad review see~\cite{jaeAnag}). A direct numerical simulation of individual grains of the substrate, or discrete element method (DEM~\cite{poschelbook}), carries with it much of the same challenges as hard ground. However, the substrate quantities of focus most relevant to questions of locomotion are macroscopic in nature. The bulk response of the granular media like the force acting on the intruder are of most import in robophysics. Grain-level details like particle orientation after a few seconds are ``washed'' out at the macroscopic level. This allows for many locomotor modes through granular media to leverage much simpler empirical force relations such as resistive force theory (RFT). Thus, despite its complexity, in many robophysics cases, granular media becomes easier to simulate than both hard ground and fluids. On the other hand, this notion is challenged during fast granular locomotion or substrates containing larger and/or complex obstacles, where the consideration of hard contacts and granular inertia (presently) necessitate the use of DEM.

\subsection{Simulating hard ground}
\label{sec:compuhard}

Modeling locomotion on hard ground, as simple as it seems, is nontrivial. When two rigid bodies collide, their velocities change according to momentum conservation and energy loss. In general, collisions are nearly instantaneous, which imposes challenges to numerical modeling (i.e., with finite time-steps). Regularization models (also known as compliance models) resolve collisions by allowing small deformations of rigid bodies in contact. The deformations, typically represented as geometric overlaps, produce spring-like forces~\cite{gonthier2004regularized}. To capture these deformations, small time steps are required. Even so, oscillations and numerical instabilities resulting from large forces are challenging to avoid with this method.

Other more state-of-the-art methods formulate the contact process into a linear/nonlinear complementarity problem (LCP/NCP). During collision, non-penetration constraints are imposed~\cite{lotstedt1982mechanical,marques1993differential,moreau1985standard} and integrated into the Newton-Euler equations of motion together with other bilateral constraints (e.g., from joints). In the absence of friction, the equations can be linear in either acceleration~\cite{baraff1993issues,pang1996complementarity} or velocity~\cite{anitescu1997formulating,stewart2000rigid,stewart1996implicit} (with proper discretization in time). Unknown constraint forces/impulses (from contacts of joints) can then be determined from optimization techniques (e.g., successive over relaxation, gradient descent). Frictional forces (Coulomb-like) introduce nonlinear constraints into the system. To re-cast as a linear complementarity problem, these forces can be discretized. However, an efficient cone constraint optimization ~\cite{anitescu2010iterative} formulation for friction has been proposed and used in Chrono::Engine, a parallel multibody dynamics engine with predictive power.

\subsection{Simulating fluids}
\label{sec:compufluid}

The partial differential equations that describe fluids (Navier-Stokes equations) are impossible to solve analytically for all but the simplest flows. Thus for a few decades, the study of aerial and aquatic locomotion (which often induces complex flows) has used numerical methods known as computational fluid dynamics (CFD). With proper boundary conditions, thrust and drag in water/air can (in principle) be solved numerically using CFD~\cite{FAUCI01121996, Fauci198885, Wolfgang01091999}. However, even in the 21st century, challenges exist with implementing CFD~\cite{jameson1996present}, such as the computational cost of achieving high order accuracy in turbulent flows and implementation of complex geometries of intruders.  In recent years, advances in parallel computing have made realistic, large-scale simulations of animal and robotic locomotion possible~\cite{Curet2010, Chen15022011, Tytell01052004, Li15112012, Miller01112012, Kern15122006, Chang05082014, Alben2013}. Stephan et al. conducted a full 3-D simulation of anguilliform swimming (relevant for eel shaped fishes) and optimized the locomotion mode~\cite{Kern15122006}. Borazjani et al. simulated the hydrodynamic body-fluid interaction of a bluegill sunfish, where the numerical flow field captured experimental PIV (particle image velocimetry) patterns~\cite{Borazjani15022012}. Wang's group developed a free flight simulation fruit fly and explored the control strategy used by the insects~\cite{Chang05082014}. The jet-propelled locomotion of jellyfish was studied by Alben et al~\cite{Alben2013}, and the results from their analytical/computational model showed good agreement with experimental data.

\subsection{Simulating granular media, DEM}
\label{sec:highspeed}

Discrete element method (DEM) computer simulations model granular media as multiple particles interacting under Newton's laws and collisional forces. DEM is simple to implement for small numbers of spherical particles. However {\em efficient} simulation of granular media is not trivial. For large scale problems like dense granular flow with millions of particles, using the LCP formulation like we do for hard ground becomes untenable. The LCP iterative solver is of order $N^2$ in time (where $N$ is the number of contacts and is proportional to the number of grains in the system) for each iteration. To make the situation more challenging, the solver needs to be called at every time step to resolve the collisions. In contrast, DEM uses compliant particles (the ``regularization method''~\cite{gonthier2004regularized}) instead. Combined with a grid partition scheme for collision detection, the time scaling can be reduced to order $~N$ for many practical problems.

The accuracy of DEM for dense granular flow is well established and reliable ~\cite{poschelbook,rapaport2004art}, provided collision force parameters (typically more than two but fewer than 6) are tuned correctly and the time step is small enough. These parameters (restitution, friction coefficients,etc.) can be empirically tuned such that simulated forces on intruders moving in a granular medium (e.g., a rod dragged horizontally) match experimental measurements~\cite{maladenIJRR,ding2011}. DEM allows one to obtain information such as forces and flow fields of the granular media that were difficult to measure in experiment. Particularly when coupled with a multi-body dynamic simulator, DEM has been useful in describing body-media interactions during locomotion, facilitating parameter variation and the development of locomotor principles~\cite{maladen2011,Maladen2011a,Qian2012,zhang2013ground}. 

To demonstrate the efficacy of this approach, we briefly discuss our work~\cite{maladenIJRR,Maladen2011a} coupling DEM to a multibody solver, Working Model 2D, to simulate both the subsurface locomotion of an undulatory robot and the animal (sandfish) that inspired the design of the robot. The simulation has two phases at each time step. In the first phase, DEM computes the interaction forces between the robot body and every particle it contacts; the state (position, velocity, orientation) of the particles are also integrated using the contact forces. In the next phase, the multibody solver updates the state of the robot, based on the constraints, controls and the accumulated interaction forces. When the contact model parameters are calibrated against physical experiments (impact and drag in the specific granular media), the combined DEM-Multibody approach was able to accurately predict the undulation efficiency of the robot for all undulation frequencies ($1-4$ Hz) and all wave amplitudes tested. The sandfish simulation also matched the animal experiment and indicated that the sandfish targets an optimal undulation strategy that maximizes its speed. 

\begin{figure}[h]
\begin{centering}
\includegraphics[width={1\hsize}]{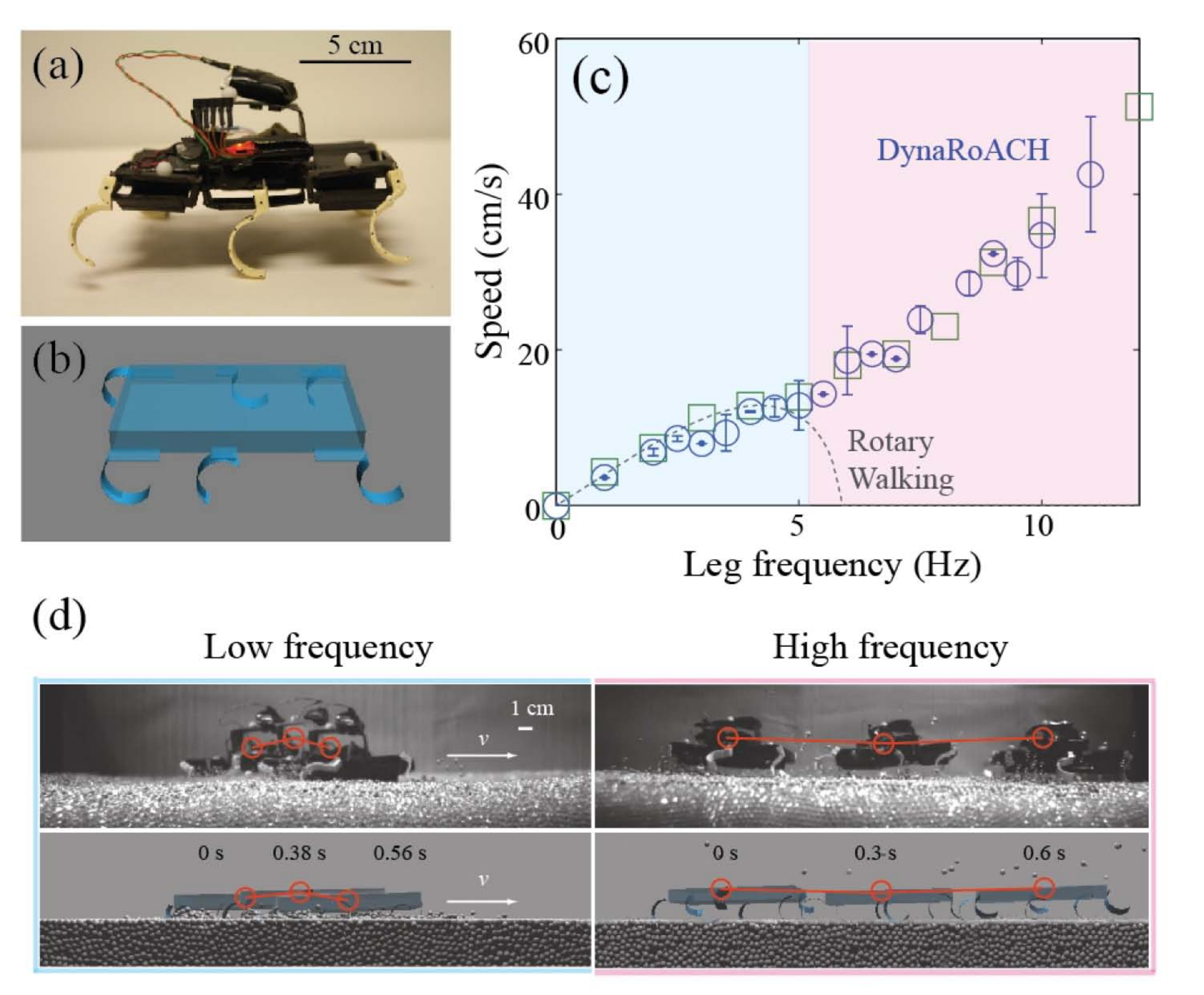}
\caption{\label{fig:lightweight} A small, lightweight robot (DynaRoACH, $10~{\rm cm}, 25~{\rm g}$), with size comparable to fast-running animals, can use two distinct propulsion mechanisms for effective locomotion on granular media: a low frequency walking mode (shaded light blue in (c)) and a high frequency running mode (shaded pink). (a) The lightweight, hexapedal DynaRoACH robot. (b) The simulated DynaRoACH robot in MBDyn~\cite{ghiringhelli1999}. (c) Average robot CoM forward speed vs. leg frequency. Blue circles represent experimental data and green circles represent simulation data. Grey dashed curve represents rotary walking model prediction (see Figure~\ref{fig:Sandbot}d) derived using a heavier robot (SandBot, $30~{\rm cm}, 2500~{\rm g}$). (d) Time sequeneces of the two locomotion modes observed for DynaRoACH moving on 3 mm glass particles. Adapted from~\cite{Qian2012}.}
\end{centering}
\end{figure}

Qian et al.~\cite{Qian2012} used DEM coupled with MBDyn~\cite{ghiringhelli1999} (a 3D multibody simulator) to investigate how a lightweight robot, the DynaRoACH~\cite{Hoover2010}, could achieve high locomotion performance on granular media. The DynaRoACH's speed in both laboratory experiments and DEM simulation agreed well, and exhibited a transition in locomotor mode from walking at low frequencies to running at high frequencies (Fig.~\ref{fig:lightweight}c). Measuring ground reaction forces in simulation (which were impossible to measure in the experiment), Qian et al. found that low frequency walking where the robot used the quasistatic ``rotary walking'' mode~\cite{Qian2015BIBM,li2009sensitive}, relied on the penetration depth-dependent hydrostatic-like forces. In contrast, high frequency gaits induced speed-dependent hydrodynamic-like forces resulting from inertial drag, allowing the robot to achieve rapid running on the leg-fluidized substrate. Zhang et al.~\cite{zhang2013ground} also demonstrated the capabilities of DEM simulation for parameter variation by varying the coefficients of particle – particle friction, particle – leg friction, and leg width over a wide range, and tested the effects of these parameters on robot locomotion performance. The particle friction parameters are difficult to vary continuously and independently of other parameters in experiment.

The future of direct granular simulations looks promising; the cone complementarity problem (CCP) formulation proposed by Tasora~\cite{tasora2011matrix} (which is also being used in Chrono::Engine) uses a matrix-free iterative solver with time proportional to N for each iteration. Compared with the DEM method, where the time step needs to be much smaller (due to grain stiffness), the non-smooth dynamic (complementarity) approach can potentially use larger time steps and remain stable. However, CCP remains under development, and substantial validation and application and results have not yet appeared.

\subsection{Modeling dry granular media, continuum descriptions}
\label{sec:compugmslow}

Continuum equations for granular media in the so-called ``rapid flow regime'' in which particles do not experience enduring contacts (i.e., the system does not exist in solid-like states) have a long history~\cite{jenAric} and have shown predictive power in shock formation in complex 3D systems~\cite{bouAmoo}. However, their efficacy has not been tested in situations relevant to locomotion, in which solid-like and fluid-like states coexist.

Thus, during such high speed locomotor interactions with granular media, which can induce complex inertial reaction forces from the substrate\cite{Katsuragi2007,Umbanhowar2010,tsiAvol05}, direct particle simulation is currently necessary. However, we have discovered that in many relevant granular locomotion scenarios that are relatively slow (low inertial number~\cite{forterre2008flows,andreotti2013granular}) we can forego DEM and, surprisingly, avoid many of the challenges of hard ground and fluid modeling. Over the past few years, our group has developed a granular resistive force theory (RFT) to describe thrust and drag forces on an intruder moving inside granular media. This approach was inspired by the early theoretical modeling of swimming of microorganisms in true fluids. In the presence of complex moving boundaries, full Navier-Stokes equations for complex flows often could not be solved (without CFD). Instead, pioneers made simpler approximations in viscous fluids. The best known of these, called Resistive Force Theory (RFT)~\cite{GRAY1955}, assumes that the deforming body can be partitioned into segments, each experiencing drag, and that the flow/force fields from these segments are hydrodynamically decoupled and do not influence the fields of other segments. Therefore, the normal and tangential forces on a small element depend only on the local properties, namely, the length of the element $ds$, the velocity $\mathbf{v}$ and the orientation $\hat{\mathbf{t}}$ (the fluid is homogeneous so position dependence is dropped). The net force on the swimmer is then computed from the integral
\begin{equation}
\mathbf{F} = \int (d\mathbf{F}_{\perp}+d\mathbf{F}_{\parallel})=\int ds (f_{\perp}(\mathbf{v},\hat{\mathbf{t}})\hat{\mathbf{n}}+f_{\parallel}(\mathbf{v},\hat{\mathbf{t}})\hat{\mathbf{t}}),
\label{eqn:RFTrelationgeneral}
\end{equation}
where the functional forms of $f_{\perp}$ and $f_{\parallel}$ can in principle be determined from the Stokes equations. The characteristics of $f_{\perp}$ and $f_{\parallel}$ are different between viscous fluids and granular media. For low speed motion in fluids, RFT forces are velocity dependant both in direction and magnitude, whereas forces in granular media are independent of the velocity magnitude. Additionally, unlike in fluids, grain-grain friction and gravity lead to a depth dependence in granular RFT forces.

To determine fluid RFT forces, experiments are often needed, because analytical solutions to the Stokes equations cannot be easily obtained even in simple cases (e.g., a finite length cylinder moving in the axial direction). Similarly in granular media, where there is no constitutive law, $f_{\perp}$ and $f_{\parallel}$ has primarily been determined from experiment or DEM simulation. Most recently in a study by Askari and Kamrin~\cite{askari2015intrusion}, however, a friction-based continuum model known as plasticity theory has been shown to reproduce experimental granular RFT measurements when simulated with FEA techniques. Moreover, the study was able to analytically uncover how, even though RFT was originally developed to simplify the analysis of viscous fluid interactions, the superposition of RFT forces is in fact more predictive in granular media than in fluids.

\begin{figure}[htbp]
\includegraphics[width=\textwidth]{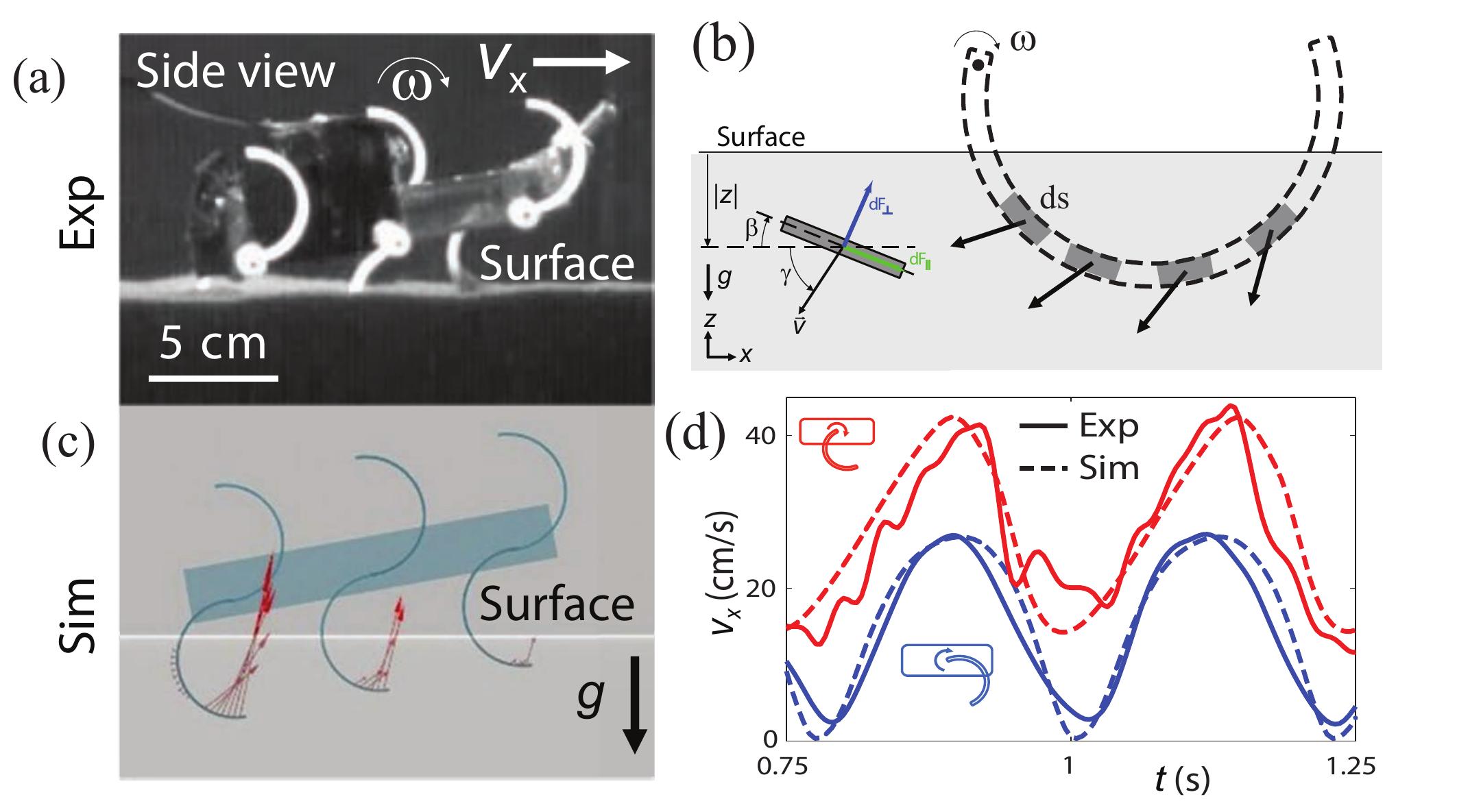}
\caption{\label{fig:figRFT} Resistive Force Theory applied to legged locomotion on dry granular media. (a) RHex-like robot (Xplorer) robot during alternating tripod gait locomotion. (b) Illustration of the basic idea behind Resistive Force Theory (RFT) for movement of a RHex c-leg into granular media. Each infinitesimal element $ds$ on the intruding leg is characterized by its tangent direction $\hat{\mathbf{t}}$ (or normal direction $\hat{\mathbf{n}}$) and its velocity $\mathbf{v}$; each element experiences a force $d\mathbf{F}_{\perp,\parallel}$. In true fluids, these forces can be described by Stokes’ law, while for granular media, they are measured in experiment. (c) Simulation of Xplorer locomotion using RFT. Red arrows indicate granular reaction forces at each segment. (d) Comparison of forward speed vs time between an experimental robot and corresponding RFT simulation using c-legs and reversed c-legs. Adapted from~\cite{li2013terradynamics} (Note: Forces were measured in $d\mathbf{F}_{z,x}$instead of $d\mathbf{F}_{\perp,\parallel}$ as shown in (b))}
\end{figure}

RFT has exhibited predictive power for legged locomotion on granular media in the quasistatic locomotor regime~\cite{li2013terradynamics, Maladen2009} (Fig.~\ref{fig:figRFT}). Using a small RHex type robot, Xplorer, the forces on the foot during slow walking modes have been found to be reproducible with the continuum equations of RFT \cite{zhang2014effectiveness}. RFT was also extensively used to investigate various aspects of sand swimming in both artifical~\cite{Maladen2011a} and biological~\cite{ding2012mechanics, sandfishtemplate} locomotors. For example, granular RFT was used to model how neuromechanical phase lag (NPL) (a phenomenon observed in many undulatory animals during which the wave of muscle activation progresses faster than the wave of body bending~\cite{sharpe2012environmental}) emerges. Ding et al.~\cite{sandfishtemplate} used RFT to explain the source of NPL during sand swimming in the frictional (non-inertial) regime. The timing of torque onset (which corresponds to muscle activation), computed from RFT, agreed well with experimentally observed electromyography signals in a sandfish lizard, indicating that NPL may depend strongly on details of the environmental interactions.

However, during fast locomotion, RFT is an insufficient descriptor of locomotion dynamics in granular media. For example, the performance of the DynaRoACH's high speed gaits~\cite{Qian2012} deviated from the rotary walking model, which, similar to RFT, assumes of quasistatic interactions. This deviation implies that hydrodynamic-like granular inertial effects contributed significantly to the reaction force, making quasistatic continuum equations like RFT ineffective.

To address if an RFT approach could be extended to the reactive force regime, we studied a jumping robot that performed a sequence of fast impulsive and forced intrusions with the objective of high speed take-off~\cite{aguilar2015jumpsand} from granular media. Non-forced impact studies suggest that such one dimensional intrusion forces can be predicted with a relation that accounts for both quasi-static frictional forces as well as velocity dependent inertial drag~\cite{Katsuragi2007}. To test these force relations, we installed the robot (originally used in hard ground jumping experiments~\cite{aguilar2012lift}) in a bed of granular media that controls volume faction (loose packed, $\phi=0.58$, to close packed, $\phi=0.63$) via air fluidization, air pulses, and bed shaking (Fig.~\ref{fig:auto}a). A separate motor raised the jumper between experiments so that the granular media's state could be reset. After hundreds of automated experimental jumps were performed, comparison between jump heights in experiment and simulation revealed that models combining inertial drag and hydrostatic friction were insufficient, inspiring a more in-depth examination of granular kinematics~\cite{aguilar2015jumpsand}.

This required a simultaneous study of both robot and substrate, which was achieved by moving the jumper next to the sidewall for direct measurement of grain kinematics via PIV analysis of high speed video capture. This analysis revealed a cone of grains that solidified underneath the foot and moved with the foot. A geometric model of this cone's growth with respect to the foot's intrusion depth in conjunction with empirical RFT measurements~\cite{li2013terradynamics} was able to predict the quasistatic resistive force on the foot. Furthermore, the cone's dynamics signaled that an added mass effect - an effect observed in fluids in which the mass of an intruder moving through a media is effectively increased as it accelerates and displaces surrounding material (see\cite{brennen1982review} for a review of added mass in fluids) - was producing a non-negligible impact on the robot's kinematics. Consequently, a 1-D reactive force theory incorporating added mass was able to reproduce experimental jump heights in simulation~\cite{aguilar2015jumpsand}.

Insight into high speed granular locomotion will improve by extending this reactive force theory, or dynamic RFT, for 3D motion. This will require a careful study of granular kinematics and dynamics during a variety of high speed locomotor interactions. We posit that it may be possible to integrate such theory with the continuum methods treating fast deforming granular media as a dense gas~\cite{jenAric}, perhaps using methods like those developed in~\cite{kamrin2012nonlocal}, which focus on steady flow. Additionally recent work by Askari and Kamrin~\cite{askari2015intrusion} suggests that a continuum model based on plasticity theory may be of use at the high speeds observed during robotic jumping.

\begin{figure}[t]
\begin{centering}
\includegraphics[width={1\hsize}]{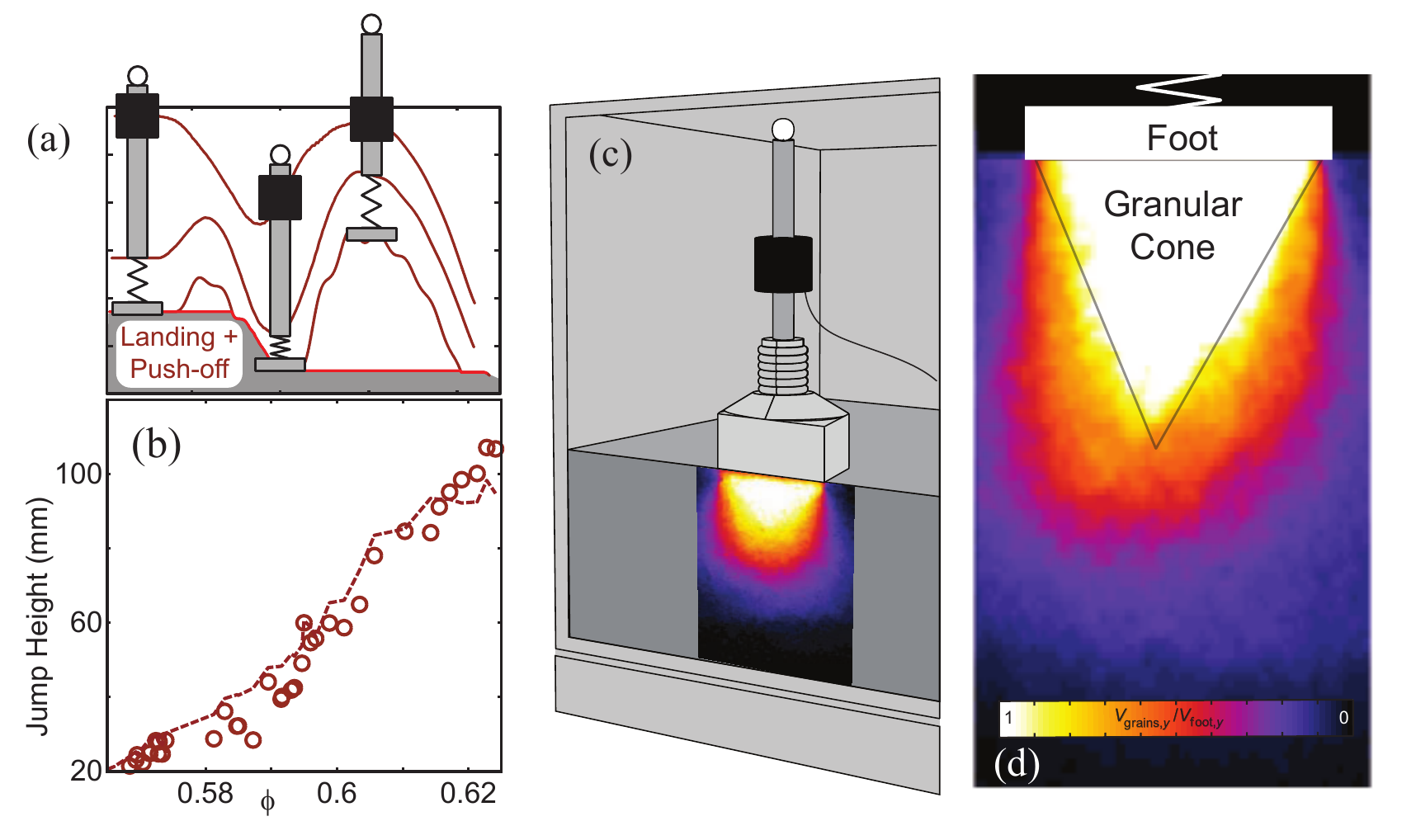}
\caption{\label{fig:sandjump} Jumping on granular media. (a) Diagram of a stutter jump in time. Motor pulls rod up for an initial hop and immediately pushes off yielding granular media upon the first landing. (b) A simulation (dashed) of the stutter jump incorporating added mass into the reaction force model of the granular media was performed using experimental forcing trajectories at different granular volume fractions and exhibited good agreement with experiment (circles). (c) This granular added mass model was intuited and confirmed from analysis of poppy seed grain kinematics during jumping, where the robot was moved up to the clear sidewall and highspeed video captured grain flow during jumping. (d) PIV velocity field of sidewall grain flow reveals a cone of material that jams under the foot and effectively makes the foot more massive, producing an added mass effect. Figure adapted from~\cite{aguilar2015jumpsand}.}
\end{centering}
\end{figure}

\section{Geometric mechanics: a language describing the building blocks for locomotion robophysics}
\label{sec:geomech}

The computational tools above provide detailed modeling capability, but do not necessarily give insight into the character or principles of locomotion. In even few degree of freedom systems (e.g., those that can be described using templates) it is difficult to develop an intuitive understanding of motion because the mapping from inputs, e.g., the movement of a joint, to outputs, e.g., net motion of center of mass, is generally complicated and nonlinear.  Recently, our group has discovered that during granular swimming (where body and material inertia are small relative to dissipative forces) we can apply tools from a discipline known as {\em geometric mechanics} (an area of research combining classical mechanics with differential geometry) to understand how shape changes (``self-deformations'') relate to translations and rotations of the body. In this section we will give a little background on these methods as applied to locomotion, arguing that such tools (which were originated by physicists) should be embraced by those interested in robophysics as a general framework and language of locomotion. For an excellent text discussing the mathematical methods that are used in this field, see~\cite{schutz1980geometrical}. For an interesting discussion of the history of these ideas (and how they appear in various contexts in classical and quantum mechanical systems) see~\cite{batterman2003falling,marsdenreviewgeo}. And for a nice introduction and discussion to geometric methods in locomotion, see~\cite{Kelly:1995}.

\begin{figure}[]
{\centering
\includegraphics[width=\textwidth]{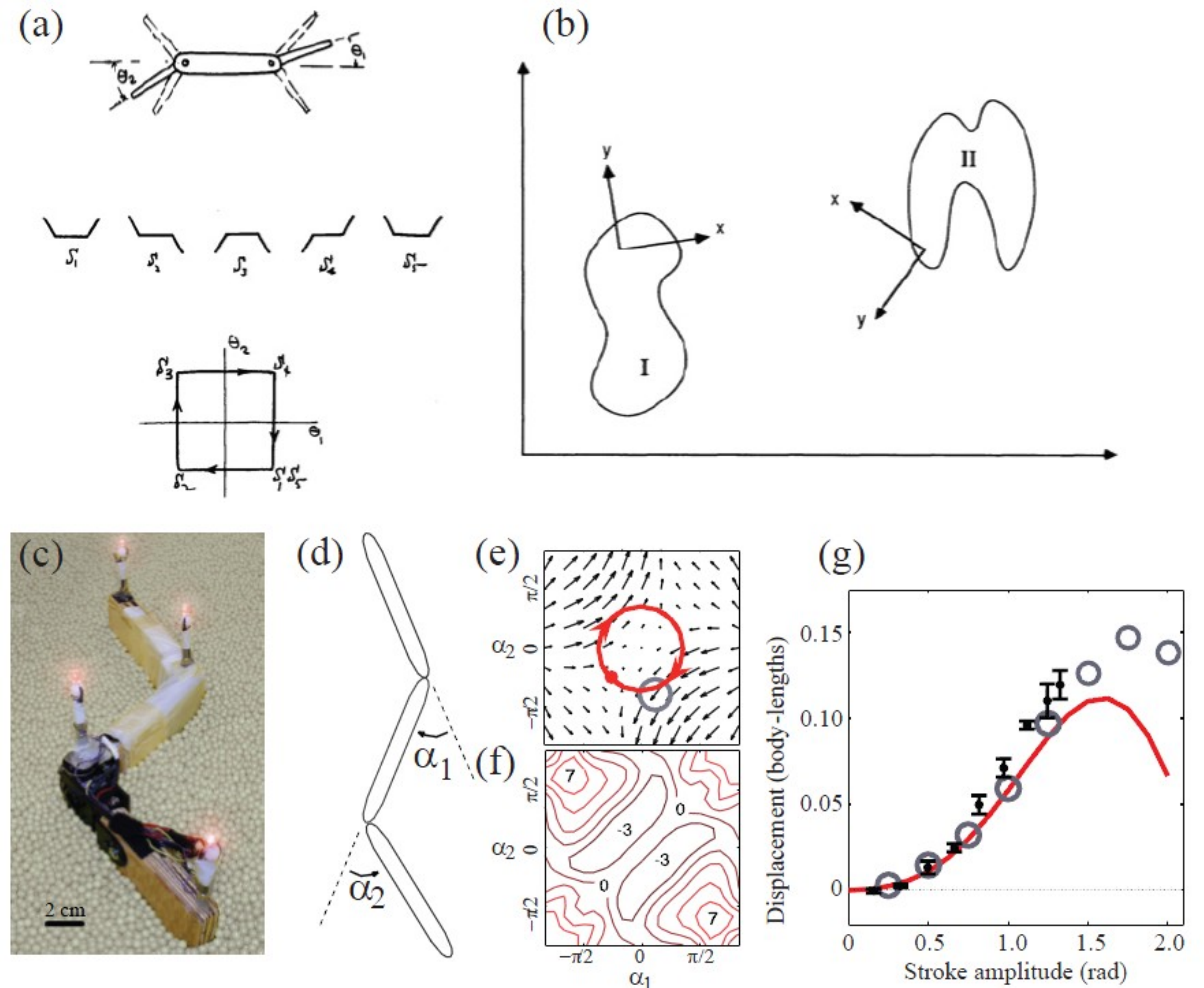}
\caption{\label{fig:geomech}
Principles of geometric mechanics, a proposed language for locomotion in complex substrates. (a) The genesis of this approach, the Purcell three-link swimmer, from~\cite{purcell1977life} and a square gait path in shape space (described by two joint angles). (b) A figure from the work of Shapere and Wilczek ~\cite{shapere1987self} who stated ``In order to measure distances between different shapes, an arbitrary choice of reference frame must be made.'' (c-g) Geometric mechanics applied to a 3-link swimmer swimming in 6 mm plastic particles. (c) Robotic 3-link swimmer. (d) Diagram of swimmer and its degrees of motion, $\alpha_1$ and $\alpha_2$. (e) Circular gait path in shape space. (f) Forward displacement curvature constraint function. (g) Forward body displacement predicted from geometric mechanic theory (red curve) compared with experiment results (dots with error bars) and DEM simulations (open circles). (c-g) adapted from~\cite{hatton2013geometric}.}
}
\end{figure}

The foundation of geometric mechanics can be found in work of physicists Shapere and Wilczek \cite{shapere1987self,shapere1989geometry}, who (inspired by Purcell's insights, discussed in ``Life at Low Reynolds Number''~\cite{purcell1977life}, see Fig.~\ref{fig:geomech}a) were the first to realize that concepts from classical mechanics, field theory, and quantum mechanics could be used to describe locomotion of microorganisms in viscous environments (see Fig.~\ref{fig:geomech}b).  Their fundamental advance was the discovery that the structure of the configuration spaces of several example systems allowed concepts from gauge theory (a type of field theory) to be applied and pointed to the existence of a mathematical framework for understanding the characeter of locomotion. This structure is intimately related to the concept of geometric phase or ``anholonomy''~\cite{batterman2003falling,wilczek1989geometric,berry1990anticipations}  In the years since, a number of researchers have contributed the field of geometric mechanics to extend its applicability to a wide range of locomoting systems.  The work of Murray and Sastry \cite{murray1993}, Ostrowski and Burdick \cite{ostrowski_98}, and Kelly and Murray \cite{Kelly:1995}, to name a few, has both modernized the language of geometric mechanics as well as generalized the scope of the field, including the analysis of systems which have nontrivial momentum evolution.

However, there have been few ~\cite{Hatton:2011IJRR,bullo2001kinematic,hatton2013geometric,shammas2007geometric} experimental tests of these theories, as early work in geometric mechanics were conducted in a purely mathematical context. Save for some early physical robotic implementations~\cite{bullo2001kinematic,shammas2007geometric}, research interests largely centered on investigating how differential equations evolve given cyclic inputs~\cite{Hatton:2011IJRR}, with a focus on evaluating idealized simple models which more readily facilitated geometric analysis. Members of our group later demonstrated that the dynamics of locomoting systems in non-ideal conditions can have mathematical structures similar to those of the ideal systems studied in the geometric mechanics literature, and that their differential equations can be populated empirically or from simulation ~\cite{hatton2013geometric}. This observation let us separate geometric gait analysis from model construction, and now makes physical experiments much more viable. We will amplify on this study in more detail below.

A key product of the theoretical work above was the development of the {\em reconstruction equation} for locomoting systems, which relates internal shape changes to the velocity of the body relative to an inertial frame.  The  reconstruction equation has been derived for a diverse set of systems, including those that locomote across land, swimming in a variety of fluid regimes, or float freely in space. The value of the reconstruction equation (and more generally in deriving geometric models), whether analytic or empirical, is that it becomes relatively straightforward to qualitatively understand the {\em character} of locomotion, without having to necessarily think about the forces that ultimately govern particular behaviors.  These methods also allow quantitative modeling of locomotor behaviors: For example, \emph{Lie bracket averaging}~\cite{murray1993} integrates the curvature of the local connection matrix $ \mathbf{A}$ to approximate the net displacement a system achieves over cyclic shape oscillations, i.e., gaits, which are defined to be closed regions of the shape space.  This displacement is referred to in the literature as geometric phase.

A particular class of systems, kinematic locomoting systems like the Purcell 3-link swimmer (Fig.~\ref{fig:geomech}a) and our sand-swimming robots (Fig.~\ref{fig:geomech}c), have proven quite amenable to study using geometric phase. Such systems have the property that the net displacement of the locomotor is a function only of the deformation and is independent of its rate.  In these systems, where displacement is either dominated by dissipative forces or, on the other side of the spectrum, momentum conservation laws, the researchers above have shown that net displacement can be approximated using a linear relationship which relates changes of shape, i.e., the system's internally controlled configuration variables, to displacement of the body.  This linear relationship between shape and body velocities is referred to as the \emph{kinematic} reconstruction equation, which has the form $\boldsymbol{\xi}= \mathbf{A}(\boldsymbol{\alpha})\cdot\boldsymbol{\dot{\alpha}}$, where $\boldsymbol{\xi}$ is the body velocity relative to the inertial frame, $\boldsymbol{\dot{\alpha}}$ is the shape velocity, and $\mathbf{A}$, the \emph{local connection}, encodes the constraints between changes in shape and changes in position. 

Choset's group has recently advanced the geometric mechanics literature for kinematic systems in two ways, both which have made it possible to apply these techniques to robophysical sand-swimming. The first is that they have developed a new method for deriving the kinematic reconstruction equation for a three-link Purcell type swimming model (Fig.~\ref{fig:geomech}d) embedded in a granular substrate.  For many different systems it is possible to analytically derive the local connection matrix (vector field), but doing so requires a linear mathematical model that describes the interaction between the system and surrounding environment.  When the environmental model is complicated, such as the models for granular media, it may be difficult, if not impossible, to derive an analytical representation of the local connection. In \cite{HattonClawar:2012,hatton2013geometric}, the authors showed that the local connection of the three-link swimmer in granular media could be derived empirically by locally perturbing its joint angles at different locations in the shape space using RFT simulation. 

The second advance has come from identification of coordinate choices that extend the quantitative accuracy of the geometric methods to large amplitude gaits. The original work in~\cite{shapere1989geometry} was developed to study movement resulting from infinitesimally small self-deformations. Historically, the approximations produced by Lie bracket averaging were regarded as quantitatively accurate for small shape oscillations but only qualitatively accurate for large gaits. The work in~\cite{Hatton:2011IJRR} allowed the construction of scalar fields called {\em Constraint Curvature Functions} (CCFs) from the local connection vector fields for large amplitude self-deformation	. The CCFs are closely related to the curls of the connection vector fields (for details see \cite{Hatton:2011IJRR}); the beauty of these structures is that net displacements induced by cyclical shape changes can be calculated as area integrals of the CCFs. Thus, it becomes relatively straightforward to both compute and visualize locomotor behaviors that result from body self-deformations.

Combining these advances has allowed our group to make the first use of geometric methods to understand locomotion in a complex medium (Fig.~\ref{fig:geomech} (c-g)), including prediction of optimal gaits and creation of novel behaviors like turning-in-place~\cite{hatton2013geometric}. For example, the contour plot in Fig.~\ref{fig:geomech}f represents the forward motion component of the curvature of $ \mathbf{A}$ for a three-link swimmer at low Reynolds number~\cite{hatton2013geometric} with joint angles $\alpha_{1}$ and $\alpha_{2}$. For the gait in Fig.~\ref{fig:geomech}e, the area integral of the CCF over the enclosed region approximates the system's net forward motion~\cite{Hatton:2011IJRR,Hatton:2011RSS}. Following a circular gait in the connection vector field in Fig.~\ref{fig:geomech}e generates a forward displacement. Computing the (signed) area of the circular gait in the CCF allows us to {\em compute} the net displacement. Increasing the radius of our circular gait reveals a peak displacement followed by a reduction in performance (Fig.~\ref{fig:geomech}g) as the gait encloses large regions of negative movement. Since this displacement is related to the area of the circle, it should scale quadratically with maximum joint angle; the experimental (and DEM simulation) data follow this trend until the circle begins to enclose regions of the CCF which are of opposite sign (lighter red regions in the CCF in Fig.~\ref{fig:geomech}f).  Thus the CCF also allows us to easily calculate the gait which yields optimal movement--we simply follow the zero-sets of the CCF in Fig.~\ref{fig:geomech}f. This produces a ``butterfly'' gait which outperforms circular gaits~\cite{hatton2013geometric}.

We are excited by the predictive and explanatory ability of geometric mechanics applied to a real-world physical system, and propose that this framework can (and should) form an essential ingredient in understanding and characterizing locomotion robophysics movement principles.  Perhaps somewhat provocatively, we speculate that (at least in non-inertial systems) appropriate CCFs  can be used in a manner analogous to free energy functions in equilibrium systems: that is, integrated areas in the CCFs provide an ``answer'' to a locomotion question, without having to compute dynamics (analogous to how free energies provide the equilibrium state of a thermodynamic system, without having to compute dynamics). Thus, geometric mechanics gives us the ability to rationally search for kinematic control templates and, in some cases, the means with which to modulate them. Aside from this speculation we anticipate that a useful future direction will be to leverage the benefits of geometric mechanics to provide insight into hybrid dynamical scenarios where locomotors exhibit transitions between aerial free fall and substrate interaction. Such situations are common during modes of locomotion like walking, running and hopping. Preliminary work indicates the efficacy of this approach in understanding movement in our flipper-driven robotic systems~\cite{McInroeMuddy} in Fig.~\ref{fig:muddy}, and in biological and robotic sidewinding.


\begin{figure}[h]
\begin{shaded}
\textbf{\begin{large}
Sidebar: Robophysics aiding biology
\end{large}}\\

\begin{centering}
\includegraphics[width={1\hsize}]{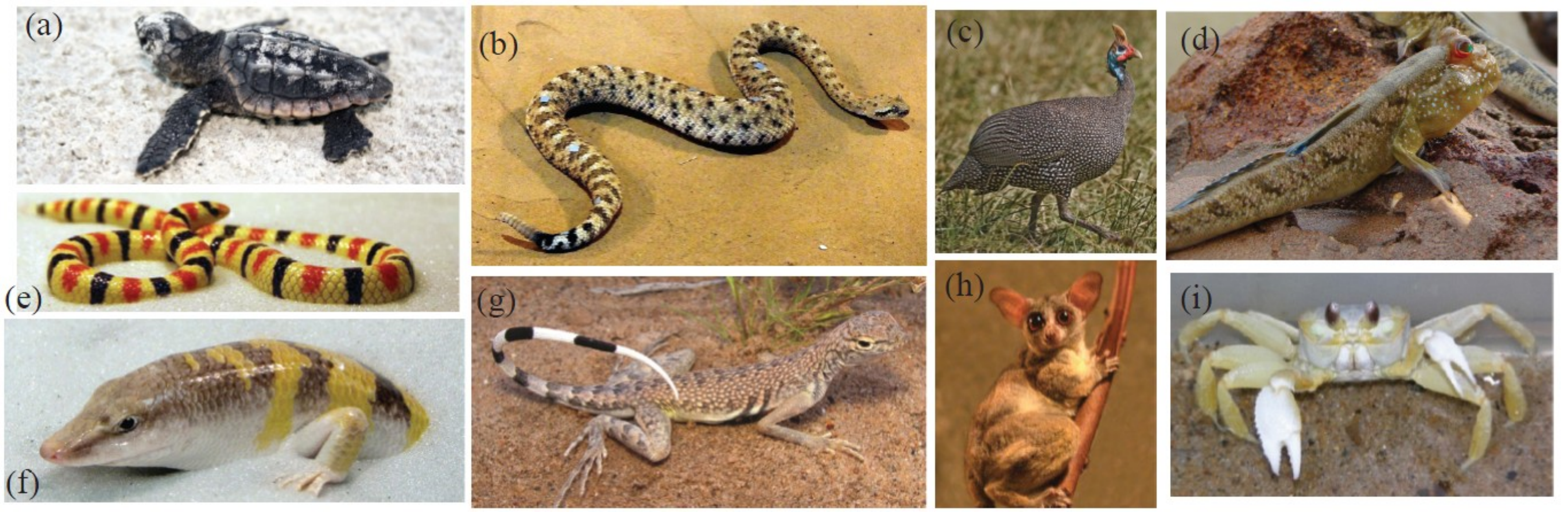}
\caption{\label{fig:animal} \scriptsize{Animals whose biomechanics are better understood as a result of robophysics. (a) Hatchling sea turtle. (b) Sidewinding rattlesnake. (c) Guineafowl. (d) Mudskipper. (e) Shovel-nose snake. (f) Sandfish lizard. (g) Zebra-tailed lizard. (h) Bushbaby. (i) Ghost crab.}}
\end{centering}

\bigskip

\scriptsize{
A careful study of direct measurements (e.g., electromyography, kinematic tracking, force measurements, post mortem as well for muscle and skeletal properties and segmental mass) of animal movement is both beneficial and often crucial to the understanding of biological locomotion. However, exclusively observing organisms can often present a variety of challenges that limit insight into the dynamical mechanisms of optimal gait performance. First, large multicellular organisms such as vertebrates are highly complex systems which regulate many functions aside from movement, often making it difficult to tease out which material, structural, mechanosensory and control features are most important for optimal performance during specific modes of locomotion. Also, the pace of research can often be limited by animal temperament. Even with proper protocols in place to ensure healthy, well fed and well rested animals that are properly motivated to perform certain locomotor tasks, organisms can often exhibit a certain degree of unpredictability in their behavior. In a laboratory setting, one must ensure that artificial environments to do not alter behavior in undesirable ways (although such changes can also be interesting to study). Aside from comparing with similar species, it can be difficult to know whether an animal is on average operating with near-optimal gait kinematics, since optimal gaits cannot be quantitatively identified with precision without systematic variation of actuation parameters. Such a variation can illuminate not only optimal movements, but also gaits that produce poor performance or failures, the conditions for which can be just as useful to identify. While animals can exhibit poor performance and failures, such situations can be difficult to systematically study (aside from restricting appendages or other methods which may also alter behavior in unpredictable ways).}\\

\scriptsize{
It is for the aforementioned reasons that robophysics is useful to aid in understanding biological locomotion. Such  models can be explicitly controlled to perform systematically varied movements to not only compare with animal performance, but also reveal the underlying physics principles of movement. Robophysics studies can give insight into the conditions for optimal performance and failure as well as the mechanical, geometric and actuation features most important in locomotion. For example, the study of a seven link sand swimmer has revealed that the mechanism of increased forward speed resulting from increase undulation frequency in sandfish is a combination of local fluidization of granular media and its resistive frictional properties~\cite{maladen2011}. The Flipperbot was used to reveal the importance of flipper rigidity and wrist bending in hatchling sea turtles to solidify sand and transition from their land-based nest environment to the ocean~\cite{mazouchova2013flipper}. Experimentation on a jumping robot designed with a minimal model describing animal hopping revealed the non-trivial importance of resonance in the proper timing of jumps as well as the emergence of an interesting jumping strategy (the stutter jump)~\cite{aguilar2012lift} observed in nature~\cite{gunther1984biomechanical}. Leg penetration ratio important for lizards geckos and crabs running on sand~\cite{Qian2015BIBM}; lighter animals can leverage inertial drag for high speed running. Robophysical experiments and models have also helped illuminate principles of animal locomotion in wetter environments. RoboClam has helped researchers to identify and test the concept of localized fluidization, which reduces the energy cost during the burrowing of razor clams in cohesive granular soils~\cite{winter2014razor}. A bioinspired robotic foil has proven ideal to allow for systematic parameter variation during undulatory swimming~\cite{lauder2011robotic}, while the robotic knife fish demonstrates the mechanics and maneuverability of counter-propagating waves actuating elongated fins~\cite{Curet2010}.
}

\end{shaded}
\end{figure}

\section{Conclusions and Outlook}

\label{sec:conclusion}

We have discussed our physics approach to studying locomoting robots, using examples largely from our work which highlight key ingredients in our definition of robophysics, the the pursuit of the discovery of principles of self generated motion. These include studying simplified laboratory robots without focus on hardened field-ready devices, subjecting these models to systematic experimentation and parameter space exploration without focusing necessarily on design and controls that enable success. Although of course such methods can be (and are) applied to any types of models, we feel that robophysical study of actual physical devices (as opposed to only theory or simulation) is critical to the advance of robophysics and can contribute to new insights in dynamical systems, soft matter, biology and  engineering robotics. We reiterate that robophysics prescribes to a mixture of the Pierce and Feynman points of view  highlighted in the lead-in to our article-- neither biological systems nor complex engineered devices can be {\em understood} or advanced without building or creating models.

While we have primarily focused on locomotion in dry homogeneous granular media (which we claim is the simplest system for study of ``flowable'' terrestrial substrates), robophysics can be a catalyst for exciting questions related to interesting forcings of other types of soft matter. The variability in natural substrates is vast. Many substrates are solid but have complex geometries, some behave like shear thickening fluids, while other substrates, such as soil, are moist heterogeneous and have soft grains. Many substrates, such as snow, can have widely varying properties which change sensitively with weather conditions~\cite{staron2014nonequilibrium}. It is crucial to think about principles by which robots can control not only their own dynamics, but also that of the environments they traverse. There are numerous directions of research interest regarding locomotion in complex substrates, including downhill granular slopes (locomotor questions of interest might include adjustments in gaits, conditions for avalanching and resulting effective surface friction), cohesive materials, mud, leaf litter, heterogeneous obstacles, and transitions between different environments. Automated laboratory systems such as a trackway like SCATTER that can systematically change environment within which to perform robotic locomotion experiments will rapidly accelerate the research of soft matter properties relevant to locomotion, as properties and dynamics will be measured in more realistic situations (simultaneously as robots interact with media), complementing more standard tools such as rheometers.

We reiterate that locomotion robophysics can be a valuable tool for biology in the 21st century. It is common in synthetic biology to think of creation of new life from the ``bottom up'', that is, by engineering microorganisms at the genetic level. However, we argue that perhaps non-organic life can also be engineered from the ``top-down'', using principles discovered through robophysical study to constrain parameter space. In this regard, we are inspired by a quote from Bialek's book on biophysics~\cite{bialek2012biophysics} regarding the search for life:{\em ``More precisely, all the molecular components of life that we know about comprise one way of generating and maintaining the state that we recognize as being alive. We don't know if there are other ways, perhaps realized on other planets. This remark might once have seemed like science fiction, and perhaps it still is, but the discovery of planets orbiting distant stars has led many people to take these issues much more seriously. Designing a search for life on other planets gives us an opportunity to think more carefully about what it means to be alive.''} Echoing Feynman's quote in the introduction to the review, we argue that ``creation'' of life-like movement by discovery of robophysical principles of locomotion allows to design a search for life right here in our laboratories on Earth.

\begin{figure}[t]
\begin{centering}
\includegraphics[width={1\hsize}]{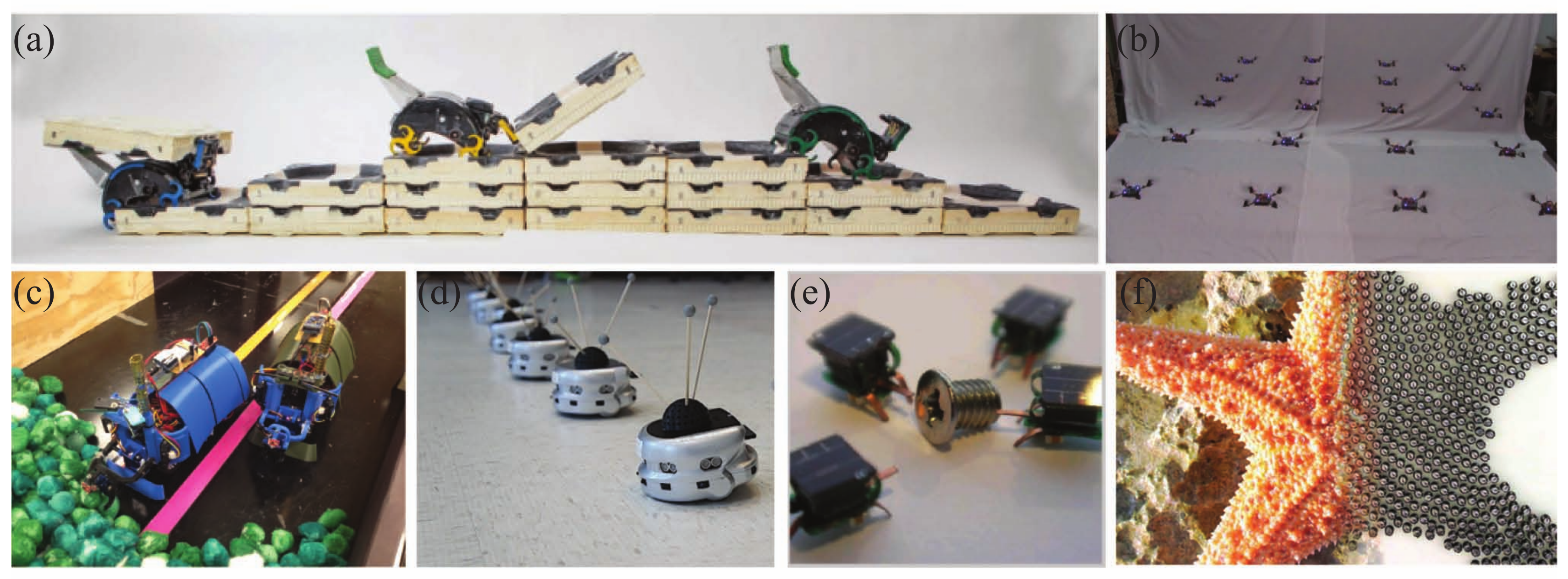}
\caption{\label{fig:collect} The current state-of-the-art in swarming robots exemplifies a field of research where rich dynamics emerge from collections of members individually following a relatively simple (though sometimes discrete, nonlinear) set of rules. The simplicity of individual behaviors has allowed for engineers to readily create swarms of robots and induce these rich dynamics without needing a full understanding of its emergence. (a) TERMES robots~\cite{werfel2014designing}. (b) GRASP micro UAVs~\cite{NM:DM:QL:10}. (c) CRABLab Ant Excavation Robots. (d) GRITSLab piano playing swarm bots~\cite{chopra2012multi}. (e) I-SWARM~\cite{ipr_1111501389}. (f) Kilobots Harvard~\cite{rubenstein2014programmable}.}
\end{centering}
\end{figure}

While our review has focused on individual locomotors moving on and within complex substrates, we cannot depart without commenting on the need for a robophysics of collective locomotion: in the future, {\em teams} of robots will build houses, dig tunnels and clean drainpipes. While there has been much work done in swarming robotics~\cite{rubenstein2014programmable,bonabeau:1999,desai:2001,ogren:2004,torney2009,saska:2014}, the typical assumption is that the agents operate in relatively sterile conditions (laboratory floors or aerial maneuvers in still air, see examples in Fig.~\ref{fig:collect}). Studying systems such as flocking robots~\cite{werfel2014designing,NM:DM:QL:10,chopra2012multi,ipr_1111501389,rubenstein2014programmable} can elucidate an understanding of how rich dynamical complexity can emerge from a collection of simplified systems. Theoretically, we expect that some of the recent hydrodynamic-like theories of active matter should be of use to understand how such collectives can accomplish tasks~\cite{berman2007algorithms}. However, when the robots come into contact with each other (in crowded conditions) or with complex substrates (digging tunnels, clambering through disaster rubble), we will need new principles to understand collective flow, clogging, etc. Consequently, we argue that parallels can be drawn between the dynamics of swarming robots and the hydrodynamics of self-propelled particles~\cite{aditi2002hydrodynamic,levine2000self,narayan2007long}. In addition, our recent biological studies of trafficking ants operating in confined conditions~\cite{gravish2015}, we have realized that physics has a role to play in such robophysics have revealed that the physics of glasses and supercooled fluids (as wells as jammed solids) can play important roles in collective movement and tasks~\cite{werfel2014designing}.

Finally, and expanding on the point above, we acknowledge that we have kept our focus on robophysics of {\em locomotion}, largely because this is our area of expertise, and because so many robots will need to move effectively in environments that can only be understood with the physics principles approach. That said, we can imagine robophysics as a more broad discipline, related to ``cybernetics'' promoted by Wiener~\cite{wienerbook} over 60 years ago. Robotic locomotors which can coexist alongside animal counterparts can fulfill the dream of one of the early cyberneticians, Louis Couffignal, who defined the field as  ``the art of ensuring the efficacy of action''~\cite{coufarticle}. However, while cybernetics became an awkward (and often maligned) relative of computer science, we feel that robophysics can fulfill the original spirit of the discipline, in that it emphasizes and adds locomotion (and to physics) to the important aspects of ``control and communication in the organism and machine'' (see also~\cite{cowan2014feedback,roth2014comparative} for recent discussion of similar issues).  Importantly (for both physics and cybernetics) robophysics emphasizes the need to do high quality, systematic experiments in conjunction with theory and modeling. 


\section{Acknowledgments}
We would like to thank NSF (PoLS, NRI, CMMI), ARL MAST CTA, ARL RCTA, ARO, DARPA, and the Burroughs Wellcome Fund for supporting this research. We would like to thank Aaron Ames, Henry Astley, Bill Bialek, Krastan Blagoev, Martin Buehler, Noah Cowan, Jin Dai, Tom Daniel, Yang Ding, Max Donelan, Robert Dudley, Ron Fearing, Bob Full, Jerry Gollub, Nick Gravish, Heinrich Jaeger, Leo Kadanoff, Dan Koditschek, George Lauder, Vadim Linevich, Malcolm MacIver, Gill Pratt, Al Rizzi, Andy Ruina, Sarah Sharpe, Sam Stanton, Harry Swinney, Joshua Weitz, Kurt Wiesenfeld and Andy Zangwill for helpful discussions over the years. Some of the thinking on this work was inspired by an NSF/ARO workshop on locomotion. HC would like to thank Joel Burdick for starting his movement towards robotics. DIG also wants to acknowledge his undergraduate research mentor Prof. John G. King (1925-2014) who instilled in him the following ideals concerning the process of experimentation: ``Make mistakes quickly'' and ``There's no substitute for hands-on fooling with real stuff.''~\cite{king2001observation}. Robophysics thrives on these philosophies.


\section{Appendix: List of Abbreviations and Acronyms}
\begin{tabular}{||c||c||}
\hline CCF & Curvature Constraint Function \\ 
\hline CFD & Computation Fluid Dynamics \\ 
\hline CPG & Central Pattern Generator \\ 
\hline DEM & Discrete Element Method \\ 
\hline GM & Granular Media \\ 
\hline PCA & Principal Component Analysis \\ 
\hline RFT & Resistive Force Theory \\ 
\hline DOF & Degrees of Freedom \\ 
\hline SCATTER & Systematic Creation of Arbitrary Terrain and Testing of Exploratory Robots \\ 
\hline SLIP & Spring Loaded Inverted Pendulum \\ 
\hline 
\end{tabular} 

\end{document}